\def\year{2018}\relax
\documentclass[letterpaper]{article}
\usepackage{aaai18}
\usepackage{times}
\usepackage{helvet}
\usepackage{courier}
\usepackage{url}  

\usepackage{amssymb}
\usepackage{graphicx}
\usepackage{epstopdf}
\usepackage{amsmath}
\usepackage{cleveref}
\crefformat{footnote}{#2\footnotemark[#1]#3}
\usepackage[ruled,vlined]{algorithm2e}
\newtheorem{lem}{Lemma}
\newtheorem{thm}{Theorem}
\newtheorem{prop}{Proposition}
\newtheorem{defn}{Definition}

\def\myproof{1}
\if\myproof1\nocopyright\fi

\begin{document}

\title{Gaussian Process Decentralized Data Fusion Meets Transfer Learning in Large-Scale Distributed Cooperative Perception}

\author{
Ruofei Ouyang \and Kian Hsiang Low\\
Department of Computer Science\\ 
National University of Singapore, Republic of Singapore\\
\{ouyang, lowkh\}@comp.nus.edu.sg 
}

\maketitle

\begin{abstract}
This paper presents novel Gaussian process decentralized data fusion algorithms exploiting the notion of agent-centric support sets for distributed cooperative perception of large-scale environmental phenomena. To overcome the limitations of scale in existing works, our proposed algorithms allow every mobile sensing agent to choose a different support set and dynamically switch to another during execution for encapsulating its own data into a local summary that, perhaps surprisingly, can still be assimilated with the other agents' local summaries (i.e., based on their current choices of support sets) into a globally consistent summary to be used for predicting the phenomenon.
To achieve this, we propose a novel transfer learning mechanism
for a team of agents capable of sharing and  transferring  information encapsulated in a summary based on a support set to that utilizing a different support set with some loss that can be theoretically bounded and analyzed.
To alleviate the issue of information loss accumulating over multiple instances of transfer learning, we propose a new information sharing mechanism to be incorporated into our  algorithms in order to achieve memory-efficient lazy transfer learning.
Empirical evaluation on real-world datasets show that our algorithms outperform the state-of-the-art methods.
\end{abstract}

\section{Introduction}
\label{sec:intro}
Central to many environmental sensing and monitoring applications (e.g., traffic flow and mobility demand predictions over urban road networks \cite{LowTASE15},
monitoring of ocean and freshwater phenomena \cite{LowSPIE09},
adaptive sampling and active sensing/learning~\cite{LowAAMAS13,NghiaICML14,LowAAMAS08,LowICAPS09,LowAAMAS11,LowICRA07,LowAAMAS12,LowAAMAS14,YehongAAAI16}, Bayesian optimization~\cite{ErikICML17,NghiaAAAI18,ling16}, among others)
is the need to scale up data fusion algorithms for big data because massive volumes of data/observations gathered by multiple static and/or mobile sensing agents have to be assimilated to form a globally consistent predictive belief of the environmental phenomenon of interest. A centralized approach to data fusion is ill-suited here because it suffers from poor scalability
in the data size and  a single point of failure.
To this end, decentralized data fusion algorithms such as distributed Bayesian filtering \cite{saber05a} 
and distributed regression \cite{guestrin04}
have been developed to improve scalability and robustness to failure.
\begin{figure*}
\begin{tabular}{ccccc}
  \includegraphics[height=2.53cm]{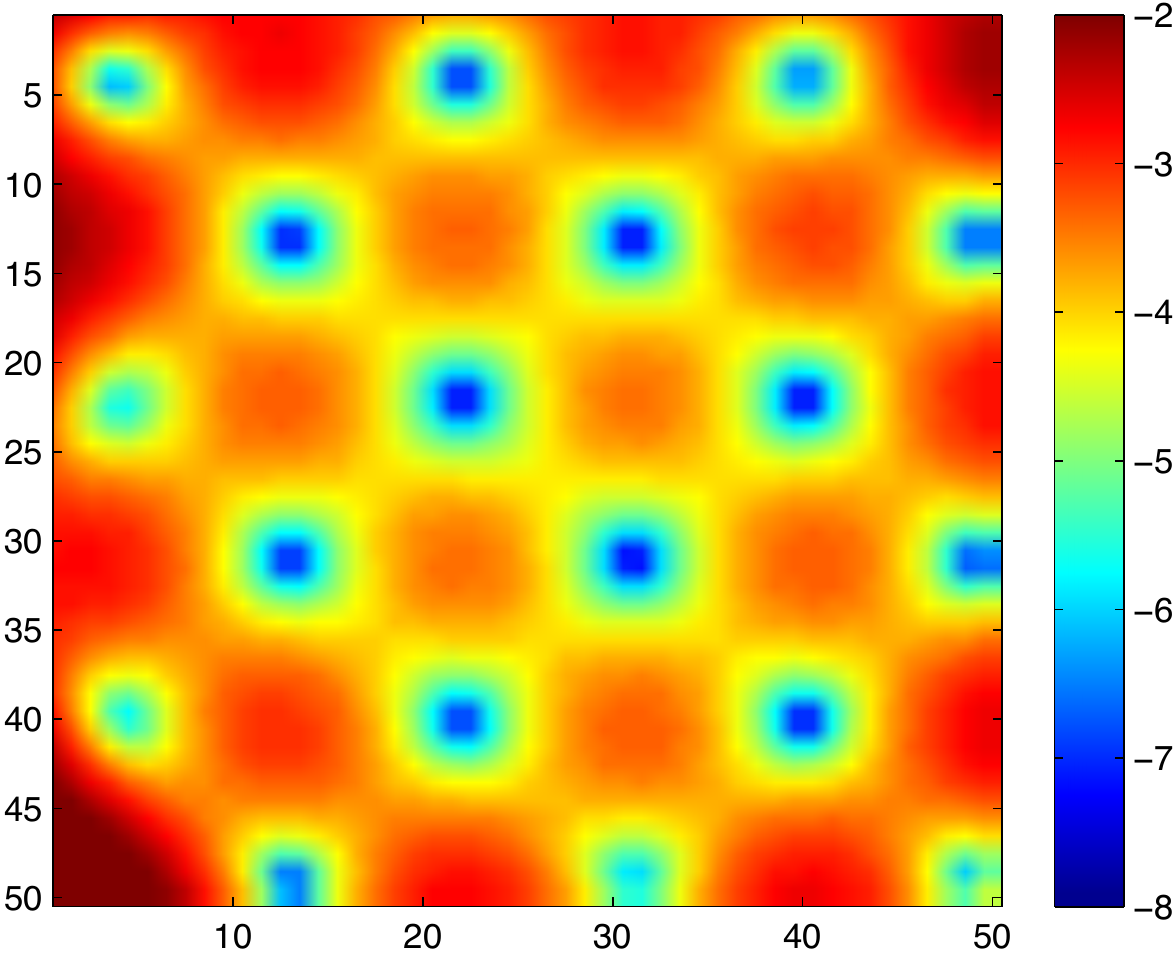} &  \includegraphics[height=2.53cm]{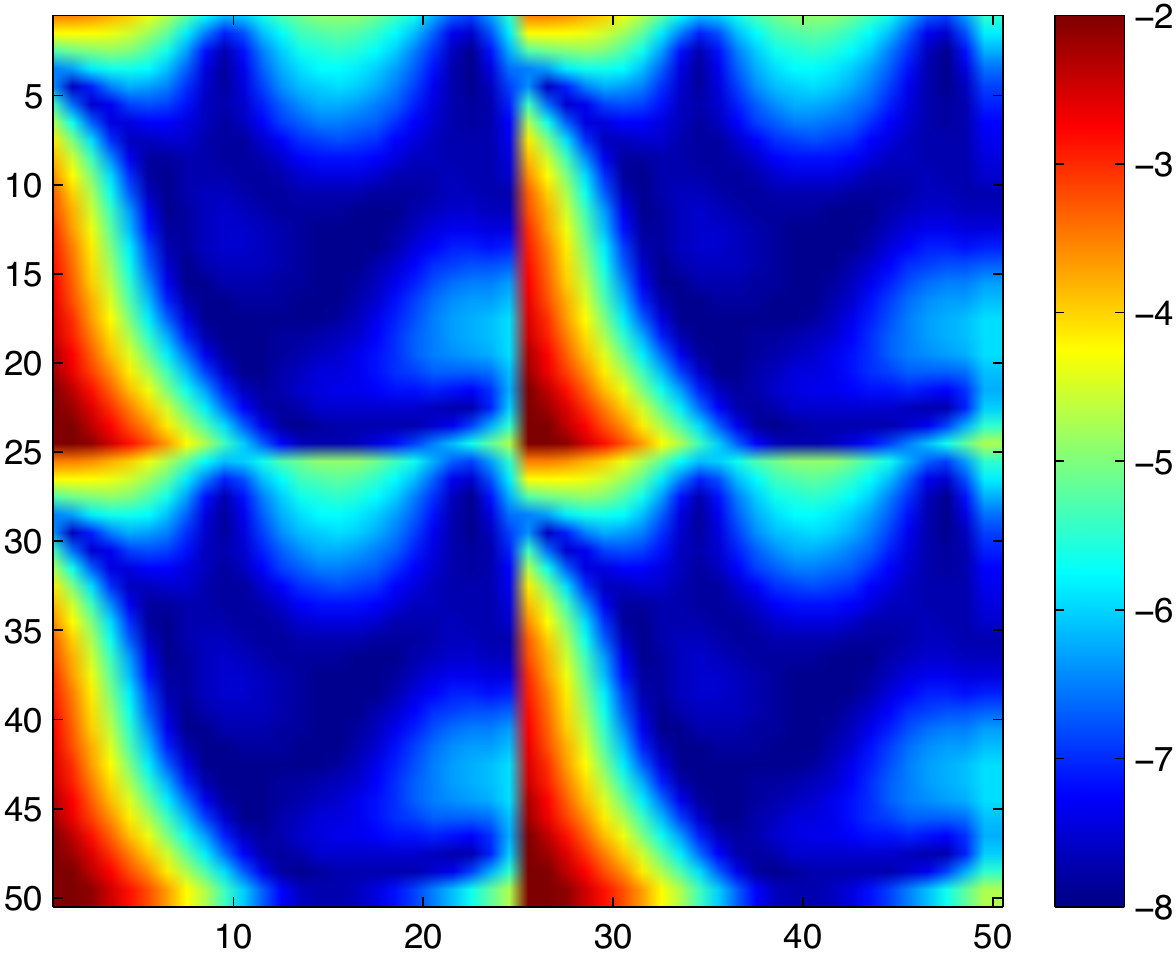} &  \includegraphics[height=2.53cm]{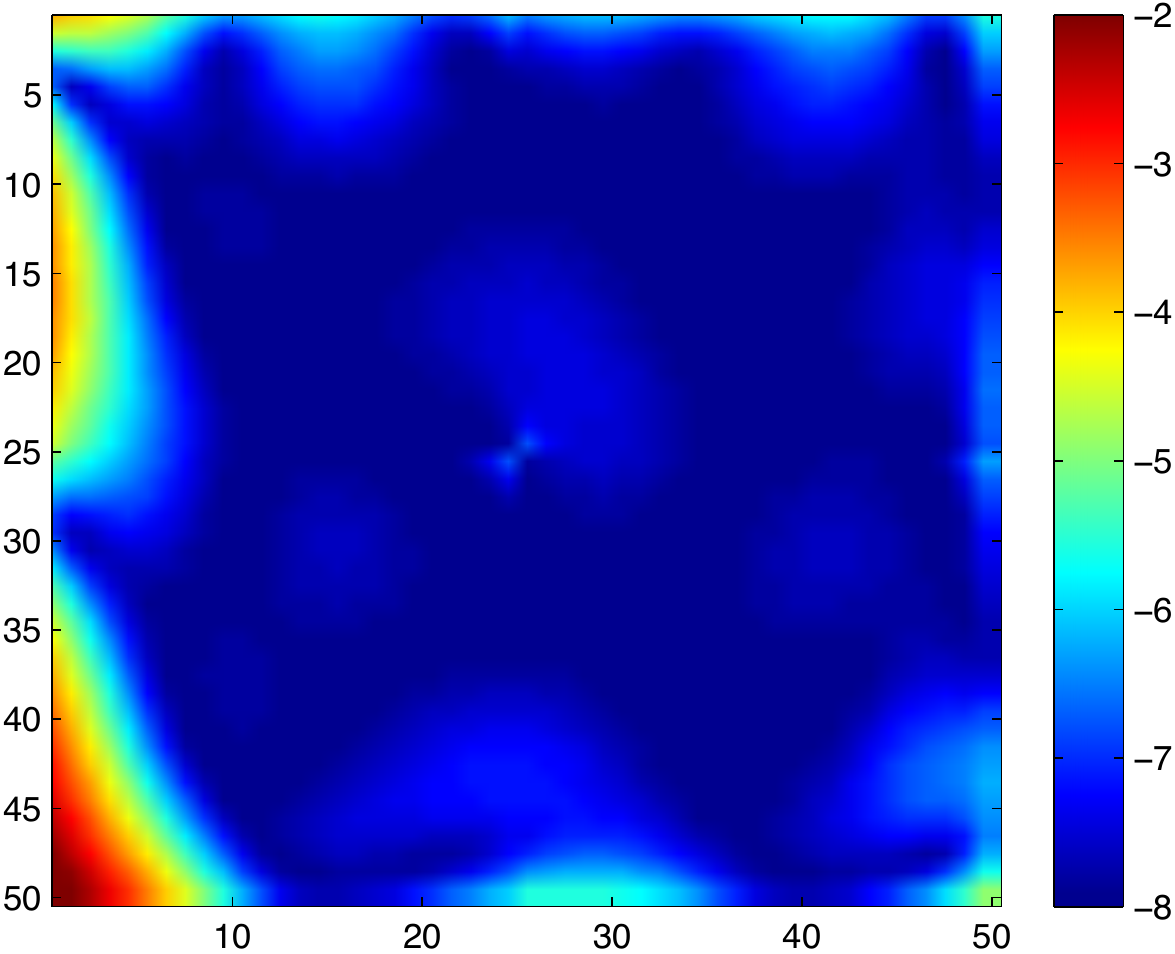} & \includegraphics[height=2.53cm]{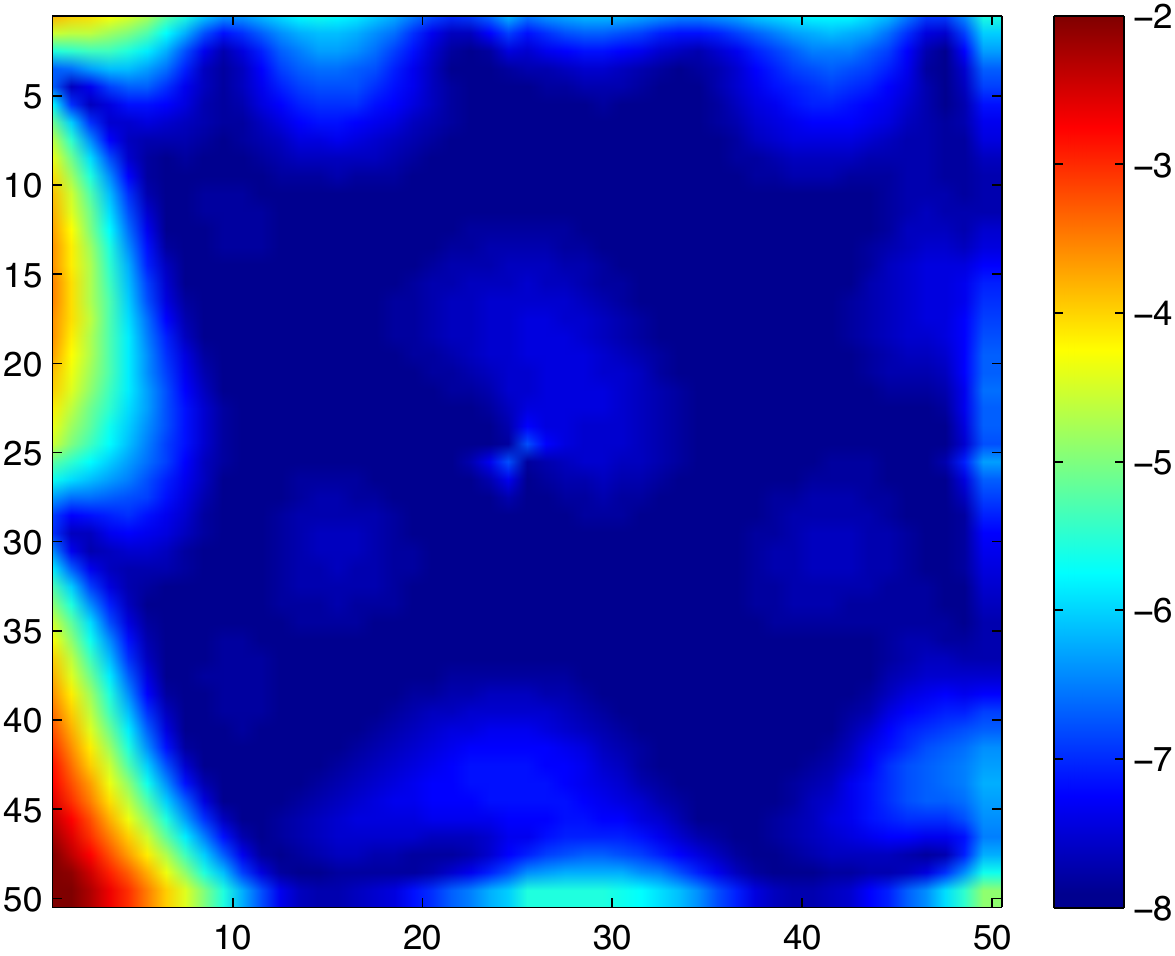} & \includegraphics[height=2.53cm]{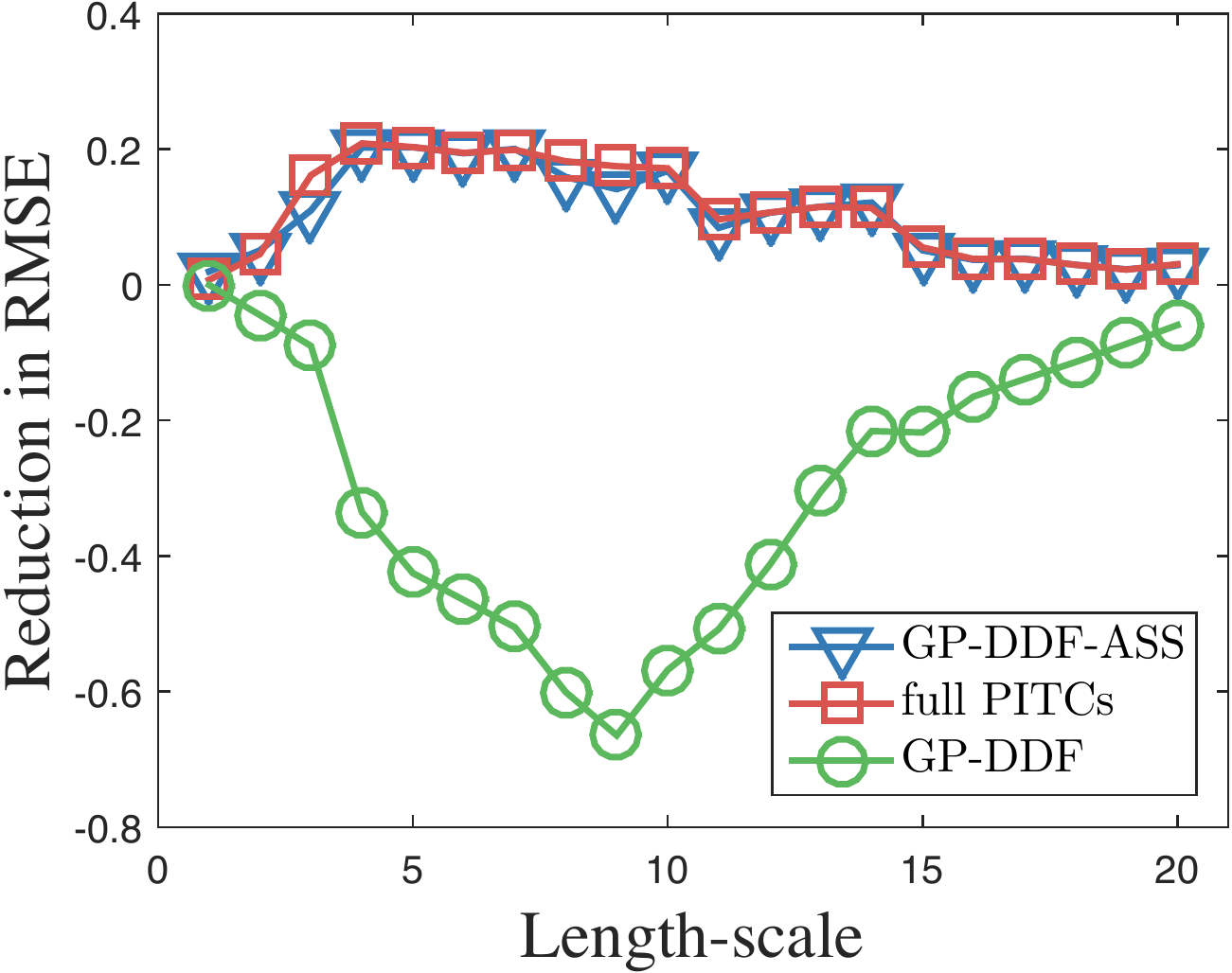} \\
 (a) GP-DDF &   (b) Local PITCs &  (c) Full PITCs & (d) GP-DDF-ASS & (e)
\end{tabular}
\caption{(a-d) Maps of log-predictive variance/uncertainty (i.e., $\log\overline{\sigma}^2_{x}$~\eqref{eqn:pred} for all $x\in\mathcal{X}$) over a simulated spatial phenomenon with length-scale of $10$ achieved by 
various decentralized data fusion algorithms given the same data and support set size for each agent, and (e) graphs of reduction in RMSE of GP-DDF, full PITCs, and GP-DDF-ASS over local PITCs vs. varying length-scales.
Experimental setup, results, and analysis for this simulated experiment are detailed in Section~\ref{toy}.}
\label{fig:demo}
\end{figure*}

Recent works \cite{LowUAI12,LowTASE15,LowRSS13,cortes09} have progressed
from the use of simple Markov parametric models assuming independent observations  (e.g., in distributed Bayesian filtering) to that of a rich class of Bayesian nonparametric \emph{Gaussian process} (GP) models characterizing continuous-valued, spatially correlated observations in order to represent the latent structure of the spatially varying,  possibly noisy phenomenon with higher fidelity.
Instead of communicating the local data of each sensing agent directly to every other agent which is not scalable, the \emph{GP decentralized data fusion} (GP-DDF) algorithms of \citeauthor{LowTASE15}~\shortcite{LowTASE15}
enable the agents to encapsulate their own data into constant-sized local summaries, exchange them, and finally assimilate them into a globally consistent summary to be exploited for predicting the phenomenon.
Different from the above distributed regression algorithms,
they do not need to exploit spatial locality assumptions for gaining efficiency and can thus be used for mobile sensing agents whose paths are not constrained by locality.
They also do not suffer from the drawbacks of the GP distributed data fusion algorithm of \citeauthor{cortes09}~\shortcite{cortes09} relying on an iterative procedure of weighted least squares, which assumes bounded correlation and uncorrelated past observations that can severely compromise its predictive performance and converges very slowly in the case of a large number of agents.
%
In contrast, the GP-DDF algorithms can be computed exactly and efficiently.
More importantly, their predictive performance can be theoretically guaranteed to be equivalent to that of sophisticated centralized sparse approximations \cite{LowUAI13,NghiaICML15,HoangICML16,MinhAAAI17,LowAAAI15,candela05,snelson07,LowAAAI14} of the GP model.

However, like their centralized counterparts, the GP-DDF algorithms rely on the notion of a \emph{fixed support set} of input locations \emph{common} to all agents for encapsulating their own data into local summaries, which raises three non-trivial issues limiting their scalability to small domains of spatial phenomena and hence small data sizes:
(a) When the domain is expanded, the support set must be increased proportionally in size to cover and predict the phenomenon well at the expense of greater time, space, and communication overheads, which grows prohibitively costly;
(b) supposing the support set is restricted in size to limit the overheads and thus only sparsely covers the large-scale phenomenon, huge information loss due to summarization (and consequently high predictive uncertainty, as shown in Fig.~\ref{fig:demo}a) is expected, especially when the local data  gathered by the possibly ``close'' agents are ``far'' (i.e., in the correlation sense) from the support set; and
(c) if the current support set needs to be replaced by a new support set of different size and input locations (e.g., due to change in domain size or time, space, and communication requirements, using an improved active learning criterion to select a support set that better covers and predicts the phenomenon),
then all  previously gathered data (if not already discarded after summarization using old support set) have to be re-encapsulated into local summaries based on the new support set, which is not scalable.

To address these challenging issues faced by GP-DDF algorithms,
this paper presents novel {Gaussian process decentralized data fusion} algorithms with \emph{agent-centric support sets}  (Section~\ref{gpddfass}) for distributed cooperative perception of large-scale environmental phenomena. In contrast to existing GP-DDF algorithms,
our proposed algorithms allow every sensing agent to choose a possibly different support set and dynamically switch to another during execution for encapsulating its own data into a local summary that, perhaps surprisingly, can still be assimilated with the other agents' local summaries (i.e., based on their current choices of support sets) into a globally consistent summary to be used for predicting the phenomenon.
To achieve this, we propose a novel \emph{transfer learning} mechanism
for a team of mobile sensing agents capable of sharing and  transferring  information encapsulated in a summary based on a support set to that utilizing a different support set with some loss that can be theoretically bounded and analyzed,
which is the \emph{main contribution} of our work here.
To alleviate the issue of information loss accumulating over multiple instances of transfer learning, we propose a new \emph{information sharing} mechanism to be incorporated into our GP-DDF algorithms with agent-centric support sets in order to achieve \emph{memory-efficient lazy transfer learning}.
As a result, our algorithms can resolve the above-mentioned critical issues plaguing existing GP-DDF algorithms: 
(a) For any unobserved input location, an agent can choose a small, constant-sized (i.e., independent of domain size of phenomenon) but sufficiently dense support set surrounding it to predict its measurement accurately with much lower predictive uncertainty (see Fig.~\ref{fig:demo}d) while preserving time, space, and communication efficiencies;
(b) the agents can reduce the information loss due to summarization by choosing or dynamically switching to a support set ``close'' to their local data;
and (c) without needing to retain previously gathered data, an agent can choose or dynamically switch to a new support set whose summary can be constructed using information transferred from the summary based on its current support set, thus preserving scalability to big data.
We empirically evaluate the performance of our algorithms using real-world datasets featuring
indoor lighting quality gathered by a team of $3$ real Pioneer $3$-DX mobile robots and sea surface temperature of the Indian ocean explored by $64$ agents; the latter is millions in size (Section~\ref{sect:expt}).
%
%
\section{Background and Notations}
\label{gpm}
\subsubsection{Modeling Spatially Varying Environmental Phenomena with Gaussian Processes (GPs).}
A GP can model a spatially varying environmental phenomenon as follows: The phenomenon is defined to vary as a realization of a GP.
Let $\mathcal{X}$ be a set representing the domain of the phenomenon such that each
 location $x\in \mathcal{X}$
is associated with a realized (random)  measurement $y_x$ ($Y_x$) if it is observed (unobserved).
Let
$\{Y_x\}_{x\in \mathcal{X}}$ denote a GP,
that is, any finite subset of $\{Y_x\}_{x\in \mathcal{X}}$ follows a multivariate Gaussian distribution. 
Then, the GP is fully specified by its \emph{prior} mean $\mu_x \triangleq \mathbb{E}[Y_x]$ and covariance $\sigma_{xx'} \triangleq \mbox{cov}[Y_x,Y_{x'}]$ for all $x, x' \in \mathcal{X}$, the latter of which characterizes the spatial correlation structure of the phenomenon and can be defined, for example, by the 
squared exponential covariance function
\begin{equation}
\sigma_{xx'} \triangleq \sigma_s^2\exp(-0.5\|{\Lambda}^{-1}({x} - {x}')\|^2) + \sigma_n^2\delta_{xx'}
\label{secf1}
\end{equation}
where $\sigma_s^2$ ($\sigma_n^2$) is its signal (noise) variance hyperparameter controlling the intensity (noise) of the measurements,
${\Lambda}$ is a diagonal matrix with length-scale hyperparameters $\ell_1$ and $\ell_2$ controlling, respectively, the degree of spatial correlation or ``similarity'' between measurements in the horizontal and vertical directions of the phenomenon,
and $\delta_{xx'}$ is a Kronecker delta that is $1$ if $x = x'$, and $0$ otherwise.

Supposing a column vector $y_\mathcal{D}\triangleq (y_{x'})^{\top}_{x'\in\mathcal{D}}$ of realized measurements is observed for some
 set $\mathcal{D} \subset \mathcal{X}$ of locations, a GP model can exploit these observations/data to perform probabilistic regression by providing a Gaussian \emph{posterior}/predictive distribution
\begin{equation}
\mathcal{N}(\mu_x+\Sigma_{x\mathcal{D}}\Sigma_{\mathcal{D}\mathcal{D}}^{-1}(y_\mathcal{D}-\mu_\mathcal{D}),\sigma_{xx}-\Sigma_{x\mathcal{D}}\Sigma_{\mathcal{D}\mathcal{D}}^{-1}\Sigma_{\mathcal{D}x})
\label{socareless}
\end{equation}
of the  measurement for any unobserved location $x \in \mathcal{X} \setminus \mathcal{D}$ where $\mu_\mathcal{D}\triangleq (\mu_{x'})^{\top}_{x'\in\mathcal{D}}$, $\Sigma_{x\mathcal{D}}\triangleq (\sigma_{xx'})_{x'\in\mathcal{D}}$,  $\Sigma_{\mathcal{D}\mathcal{D}}\triangleq (\sigma_{x'x''})_{x',x''\in\mathcal{D}}$, and $\Sigma_{\mathcal{D}x}\triangleq\Sigma_{x\mathcal{D}}^{\top}$.
To predict the phenomenon, a naive approach to data fusion is to fully communicate all the data to every mobile sensing agent, each of which then predicts the phenomenon separately using the Gaussian predictive distribution in \eqref{socareless}.
Such an approach, however, scales poorly in the data size $|\mathcal{D}|$
due to the need to invert $\Sigma_{\mathcal{D}\mathcal{D}}$ which incurs  $\mathcal{O}(|\mathcal{D}|^3)$ time.

\subsubsection{GP Decentralized Data Fusion (GP-DDF).}
\label{notcareful}
To improve the scalability of the GP model for practical use in data fusion,
the work of \citeauthor{LowTASE15}~\shortcite{LowTASE15} has proposed efficient and scalable GP decentralized data fusion algorithms for cooperative perception of environmental phenomena that can distribute the computational load among the mobile sensing agents.
The intuition of the GP-DDF algorithm of \citeauthor{LowTASE15}~\shortcite{LowTASE15} is as follows: Each of the $N$ mobile sensing agents constructs a local summary of the data/observations taken along its own path based on a common support set $\mathcal{S}\subset\mathcal{X}$ known to all the other agents and communicates its local summary to them. Then, it assimilates the local summaries received from the other agents into a globally consistent summary which is used to compute a Gaussian predictive distribution for predicting the phenomenon.
Formally, the local and global summaries and the Gaussian predictive distribution induced by GP-DDF are defined as follows:
%
%
%
\begin{defn}[Local Summary]\label{def:localsum}
Given a common support set $\mathcal{S}\subset\mathcal{X}$ known to all $N$ mobile sensing agents,
each agent $i$ encapsulates a column vector $y_{\mathcal{D}_i}$ of realized measurements for its observed locations $\mathcal{D}_i$
into a local summary $(\nu_{\mathcal{S}|\mathcal{D}_i},\Psi_{\mathcal{S}\mathcal{S}|\mathcal{D}_i})$ where
\begin{equation}
\begin{array}{rcl}
\nu_{\mathcal{B}|\mathcal{D}_i} &\triangleq &\displaystyle \Sigma_{\mathcal{B}\mathcal{D}_i}\Sigma_{\mathcal{D}_i\mathcal{D}_i|\mathcal{S}}^{-1}(y_{\mathcal{D}_i}-\mu_{\mathcal{D}_i})\ ,\\
\Psi_{\mathcal{B}\mathcal{B}'|\mathcal{D}_i} &\triangleq & \displaystyle \Sigma_{\mathcal{B}\mathcal{D}_i}\Sigma_{\mathcal{D}_i\mathcal{D}_i|\mathcal{S}}^{-1}\Sigma_{\mathcal{D}_i\mathcal{B}'}
\end{array}
\label{eqn:localsum}
\end{equation}
for all $\mathcal{B},\mathcal{B}'\subset\mathcal{X}$ and $\Sigma_{\mathcal{D}_i\mathcal{D}_i|\mathcal{S}} \triangleq \Sigma_{\mathcal{D}_i\mathcal{D}_i}-\Sigma_{\mathcal{D}_i\mathcal{S}}\Sigma_{\mathcal{S}\mathcal{S}}^{-1}\Sigma_{\mathcal{S}\mathcal{D}_i}$.
\end{defn}
%
%
\begin{defn}[Global Summary]\label{def:globalsum}
Given a common support set $\mathcal{S}\subset\mathcal{X}$ known to all $N$ mobile sensing agents and
the local summary $(\nu_{\mathcal{S}|\mathcal{D}_i},\Psi_{\mathcal{S}\mathcal{S}|\mathcal{D}_i})$ of every agent $i = 1, \ldots, N$, a global summary is defined as a tuple $(\dot{\nu}_{\mathcal{S}},\dot{\Psi}_{\mathcal{S}\mathcal{S}})$ where
\begin{equation}\label{eqn:globalsum}
\begin{array}{c}
\dot{\nu}_{\mathcal{S}} \triangleq \sum_{i=1}^N \nu_{\mathcal{S}|\mathcal{D}_i} \ , \ 
\dot{\Psi}_{\mathcal{S}\mathcal{S}} \triangleq  \sum_{i=1}^N \Psi_{\mathcal{S}\mathcal{S}|\mathcal{D}_i} + \Sigma_{\mathcal{S}\mathcal{S}}\ . 
\end{array}
\end{equation}
\end{defn}
\begin{defn}[GP-DDF]\label{def:gpddf}
Given a common support set $\mathcal{S}\subset\mathcal{X}$ known to all $N$ agents and the global summary $(\dot{\nu}_{\mathcal{S}},\dot{\Psi}_{\mathcal{S}\mathcal{S}})$, the GP-DDF algorithm run by each agent computes a  Gaussian predictive distribution $\mathcal{N}(\overline{\mu}_{x}, \overline{\sigma}^2_{x})$ of the measurement for any unobserved location $x \in \mathcal{X} \setminus \mathcal{D}$ where\vspace{-0mm}
%
%
\begin{equation}\label{eqn:pred}
\begin{array}{rcl}
\overline{\mu}_{x}&\triangleq &\mu_{x}+\Sigma_{x\mathcal{S}}\dot{\Psi}_{\mathcal{S}\mathcal{S}}^{-1}\dot{\nu}_{\mathcal{S}}\ ,\\
\overline{\sigma}^2_{x} &\triangleq &\sigma_{xx} - \Sigma_{x\mathcal{S}}(\Sigma_{\mathcal{S}\mathcal{S}}^{-1}-\dot{\Psi}_{\mathcal{S}\mathcal{S}}^{-1})\Sigma_{\mathcal{S}x}\ .
\end{array}
\end{equation}
\end{defn}
The Gaussian predictive distribution~\eqref{eqn:pred} computed by the GP-DDF algorithm
 is theoretically guaranteed by \citeauthor{LowTASE15}~\shortcite{LowTASE15} to be equivalent to that   induced by the centralized \emph{partially independent training conditional} (PITC) approximation \cite{candela05} of the GP model. Running GP-DDF  on each of the $N$ agents can, however, reduce the $\mathcal{O}(|\mathcal{D}|((|\mathcal{D}|/N)^2 + |\mathcal{S}|^2))$ time incurred by PITC  to
only $\mathcal{O}((|\mathcal{D}|/N)^3 + |\mathcal{S}|^3 + |\mathcal{S}|^2 N)$ time, hence scaling considerably better with increasing data size $|\mathcal{D}|$.
%

Though GP-DDF scales well with big data, it can predict poorly due to information loss caused by summarizing the measurements and correlation structure of the data/observations and sparse coverage of the areas with highly varying measurements by the support set. To address its shortcoming,
the GP-DDF$^+$ algorithm of \citeauthor{LowTASE15}~\shortcite{LowTASE15}
additionally exploits the data local to an agent
to improve the predictions for unobserved locations ``close'' to its data (in the correlation sense) while
preserving the efficiency of GP-DDF by adopting its idea of summarizing information into  local and global summaries (Definitions~\ref{def:localsum} and~\ref{def:globalsum}).
The Gaussian predictive distribution computed by GP-DDF$^+$\if\myproof1 (Appendix~\ref{great}) \fi\if\myproof0 \cite{Ruofei17} \fi
is theoretically guaranteed by \citeauthor{LowTASE15}~\shortcite{LowTASE15} to be equivalent to that induced by the centralized \emph{partially independent conditional} (PIC) approximation \cite{snelson07} of the GP model.
%
GP-DDF$^+$ shares the same improvement in scalability over PIC as that of GP-DDF over PITC.
%
%
%
\section{GP-DDF with Agent-Centric Support Sets}
\label{gpddfass}
{\bf Transfer Learning.}
\label{sect:tl}
It can be observed from~Section~\ref{gpm}
that the GP-DDF and GP-DDF$^+$ algorithms depend on a common support set $\mathcal{S}$ known to all $N$ mobile sensing agents, which raises three non-trivial issues previously discussed in Section~\ref{sec:intro}: (a) Their cubic time cost in $|\mathcal{S}|$ prohibits increasing the size of $\mathcal{S}$ too much to preserve their efficiency, which consequently limits the expansion of the domain of the phenomenon for which it can still be covered and predicted well;
(b) if $\mathcal{S}$ sparsely covers the large-scale phenomenon due to its restricted size and is thus ``far'' from the data and unobserved locations to be predicted, then the values of the components in terms like $\Sigma_{\mathcal{S}\mathcal{D}_i}$ and $\Sigma_{x\mathcal{S}}$ tend to zero, which degrade their predictive performance; and (c) when switching to a new support set, they have to wastefully discard all previous summaries based on the old support set.

To address the above issues, a straightforward approach inspired by the local GPs method 
is to partition the domain of the phenomenon into local areas and run GP-DDF or GP-DDF$^+$ with a different, sufficiently dense support set for each local area.
Such an approach often
suffers from discontinuities in predictions and very high predictive uncertainty at the boundaries between local areas
 (see Fig.~\ref{fig:demo}b) and
only utilizes the data within a local area for its predictions, thereby performing poorly in local areas with little/no data.
These drawbacks motivate the need to design and develop a transfer learning mechanism for a team of mobile sensing agents capable of sharing and transferring information encapsulated in a summary based on a support set for a local area to that utilizing a different support set for another area. In this section, we will describe our  novel transfer learning mechanism and its use in our GP-DDF or GP-DDF$^+$ algorithm with agent-centric support sets and theoretically bound and analyze its resulting loss of information.

Specifically, supposing a mobile sensing agent $i$ moves from a local area with support set $\mathcal{S}$ to another local area with a different support set $\mathcal{S}'$ (i.e., $\mathcal{S}\bigcap\mathcal{S}' = \emptyset$), the local summary $(\nu_{\mathcal{S}'|\mathcal{D}_i},\Psi_{\mathcal{S}'\mathcal{S}'|\mathcal{D}_i})$ based on the new support set $\mathcal{S}'$ can be derived \emph{exactly} from the local summary $(\nu_{\mathcal{S}|\mathcal{D}_i},\Psi_{\mathcal{S}\mathcal{S}|\mathcal{D}_i})$  utilizing the old support set $\mathcal{S}$ only when the data $(\mathcal{D}_i,y_{\mathcal{D}_i})$ gathered by agent $i$ (i.e., discarded after encapsulating into $(\nu_{\mathcal{S}|\mathcal{D}_i},\Psi_{\mathcal{S}\mathcal{S}|\mathcal{D}_i})$) in the local area with support set $\mathcal{S}$ can be \emph{fully} recovered from $(\nu_{\mathcal{S}|\mathcal{D}_i},\Psi_{\mathcal{S}\mathcal{S}|\mathcal{D}_i})$, which is unfortunately not possible.
Our key idea is thus to derive the local summary $(\nu_{\mathcal{S}'|\mathcal{D}_i},\Psi_{\mathcal{S}'\mathcal{S}'|\mathcal{D}_i})$ \emph{approximately} from $(\nu_{\mathcal{S}|\mathcal{D}_i},\Psi_{\mathcal{S}\mathcal{S}|\mathcal{D}_i})$
in an efficient and scalable manner by exploiting the following important definition:
\begin{defn}[Prior Summary] Given a support set $\mathcal{S}\subset\mathcal{X}$ for a local area, each mobile sensing agent $i$ encapsulates a column vector $y_{\mathcal{D}_i}$ of realized measurements for its observed locations $\mathcal{D}_i$
into a prior summary $(\omega_{\mathcal{S}|\mathcal{D}_i},\Phi_{\mathcal{S}\mathcal{S}|\mathcal{D}_i})$ where
\begin{equation}\label{eqn:priorsum}
\begin{array}{rcl}
\omega_{\mathcal{S}|\mathcal{D}_i}& \triangleq & \Sigma_{\mathcal{S}\mathcal{D}_i}\Sigma_{\mathcal{D}_i\mathcal{D}_i}^{-1}(y_{\mathcal{D}_i}\hspace{-1mm}-\mu_{\mathcal{D}_i})\ ,\\
\Phi_{\mathcal{S}\mathcal{S}|\mathcal{D}_i} &\triangleq & \Sigma_{\mathcal{S}\mathcal{D}_i}\Sigma_{\mathcal{D}_i\mathcal{D}_i}^{-1}\Sigma_{\mathcal{D}_i\mathcal{S}}\ .
\end{array}
\end{equation}
\end{defn}
The prior summary $(\omega_{\mathcal{S}|\mathcal{D}_i},\Phi_{\mathcal{S}\mathcal{S}|\mathcal{D}_i})$~\eqref{eqn:priorsum} is defined in a similar manner to the local summary $(\nu_{\mathcal{S}|\mathcal{D}_i},\Psi_{\mathcal{S}\mathcal{S}|\mathcal{D}_i})$~\eqref{eqn:localsum} except for the $\Sigma_{\mathcal{D}_i\mathcal{D}_i}$ term in the former replacing the $\Sigma_{\mathcal{D}_i\mathcal{D}_i|\mathcal{S}}$ term in the latter and is the main ingredient for making our proposed transfer learning mechanism efficient and scalable.
Interestingly, the prior summary based on the new support set $\mathcal{S}'$ can be approximated from the prior summary utilizing the old support set $S$ as follows:
\begin{prop}\label{prop0}
If $Y_{\mathcal{S}'}$ and $Y_{\mathcal{D}_i}$ are conditionally independent given $Y_{\mathcal{S}}$ (i.e., $\Sigma_{\mathcal{S}'\mathcal{D}_i|\mathcal{S}}=  \Sigma_{\mathcal{S}'\mathcal{D}_i} - \Sigma_{\mathcal{S}'\mathcal{S}}\Sigma_{\mathcal{S}\mathcal{S}}^{-1}\Sigma_{\mathcal{S}\mathcal{D}_i} =\underline{0}$) for $i=1,\ldots,N$, then
\begin{equation}
\begin{array}{rcl}
\omega_{\mathcal{S}'|\mathcal{D}_i}&=&\Sigma_{\mathcal{S}'\mathcal{S}}\Sigma_{\mathcal{S}\mathcal{S}}^{-1}\omega_{\mathcal{S}|\mathcal{D}_i} ,\\
\Phi_{\mathcal{S}'\mathcal{S}'|\mathcal{D}_i} &=& \Sigma_{\mathcal{S}'\mathcal{S}}\Sigma_{\mathcal{S}\mathcal{S}}^{-1}\Phi_{\mathcal{S}\mathcal{S}|\mathcal{D}_i}\Sigma_{\mathcal{S}\mathcal{S}}^{-1}\Sigma_{\mathcal{S}\mathcal{S}'} .
\end{array}
\label{eqn:moving}
\vspace{-0mm}
\end{equation}
\end{prop}
Its proof is in\if\myproof1 Appendix~\ref{Apx.ppt}. \fi\if\myproof0 \cite{Ruofei17}. \fi

\noindent
\emph{Remark}. The conditional independence assumption in Proposition~\ref{prop0} extends that on the training conditionals of PITC and PIC (Section~\ref{notcareful})
which have already assumed conditional independence of $Y_{\mathcal{D}_1},\ldots,Y_{\mathcal{D}_N}$ given $Y_{\mathcal{S}}$.
Alternatively, it can be interpreted as a low-rank covariance matrix approximation $\Sigma_{\mathcal{S}'\mathcal{S}}\Sigma_{\mathcal{S}\mathcal{S}}^{-1}\Sigma_{\mathcal{S}\mathcal{D}_i}$ of $\Sigma_{\mathcal{S}'\mathcal{D}_i}$.
The quality of this approximation will be theoretically guaranteed later.
%

To efficiently and scalably derive the local summary $(\nu_{\mathcal{S}'|\mathcal{D}_i},\Psi_{\mathcal{S}'\mathcal{S}'|\mathcal{D}_i})$ approximately from $(\nu_{\mathcal{S}|\mathcal{D}_i},\Psi_{\mathcal{S}\mathcal{S}|\mathcal{D}_i})$, our  transfer learning mechanism will first have to transform the local summary $(\nu_{\mathcal{S}|\mathcal{D}_i},\Psi_{\mathcal{S}\mathcal{S}|\mathcal{D}_i})$ to the prior summary $(\omega_{\mathcal{S}|\mathcal{D}_i},\Phi_{\mathcal{S}\mathcal{S}|\mathcal{D}_i})$ based on the old support set $\mathcal{S}$, then use the latter to approximate the prior summary $(\omega_{\mathcal{S}'|\mathcal{D}_i},\Phi_{\mathcal{S}'\mathcal{S}'|\mathcal{D}_i})$ based on the new support set $\mathcal{S}'$ by exploiting Proposition~\ref{prop0}, and finally transform the approximated prior summary back to approximate the local summary $(\nu_{\mathcal{S}'|\mathcal{D}_i},\Psi_{\mathcal{S}'\mathcal{S}'|\mathcal{D}_i})$, as detailed in Algorithm~\ref{alg:local} below.
The above two transformations can be achieved by establishing the following relationship between the local summary and prior summary:
\begin{prop}\label{prop1}
Given a support set $\mathcal{S}\subset\mathcal{X}$ for a local area, the local summary $(\nu_{\mathcal{S}|\mathcal{D}_i},\Psi_{\mathcal{S}\mathcal{S}|\mathcal{D}_i})$~\eqref{eqn:localsum} and the prior summary $(\omega_{\mathcal{S}|\mathcal{D}_i},\Phi_{\mathcal{S}\mathcal{S}|\mathcal{D}_i})$~\eqref{eqn:priorsum} of  agent $i$ are related by
\begin{equation}
\begin{array}{c}
\Phi_{\mathcal{S}\mathcal{S}|\mathcal{D}_i}^{-1}\omega_{\mathcal{S}|\mathcal{D}_i} \hspace{-1mm}= \hspace{-0.5mm}\Psi_{\mathcal{S}\mathcal{S}|\mathcal{D}_i}^{-1}\nu_{\mathcal{S}|\mathcal{D}_i}\ ,\
\Phi_{\mathcal{S}\mathcal{S}|\mathcal{D}_i}^{-1}\hspace{-1mm} =\hspace{-0.5mm} \Psi_{\mathcal{S}\mathcal{S}|\mathcal{D}_i}^{-1}\hspace{-1mm}+\hspace{-0.5mm}\Sigma_{\mathcal{S}\mathcal{S}}^{-1}\ .
\end{array}
\label{eqn:ppl}
\end{equation}
\end{prop}
Its proof is in\if\myproof1 Appendix~\ref{Apx.l2p}. \fi\if\myproof0 \cite{Ruofei17}. \fi

Supposing  agent $i$ has gathered additional data $(\mathcal{D}'_i, y_{\mathcal{D}'_i})$ from the  local area with the new support set $\mathcal{S}'$, it can be encapsulated into a local summary $(\nu_{\mathcal{S}'|\mathcal{D}'_i}, \Psi_{\mathcal{S}'\mathcal{S}'|\mathcal{D}'_i})$ that is assimilated with the approximated local summary  $(\nu_{\mathcal{S}'|\mathcal{D}_i},\Psi_{\mathcal{S}'\mathcal{S}'|\mathcal{D}_i})$ by simply summing them up:
\begin{equation}\label{eqn:update}
\begin{array}{rcl}
\nu_{\mathcal{S}'|\mathcal{D}_i \bigcup \mathcal{D}'_i} &=&\displaystyle \nu_{\mathcal{S}'|\mathcal{D}_i} + \nu_{\mathcal{S}'|\mathcal{D}'_i}\ ,\\
\Psi_{\mathcal{S}'\mathcal{S}'|\mathcal{D}_i \bigcup \mathcal{D}'_i}&=& \displaystyle \Psi_{\mathcal{S}'\mathcal{S}'|\mathcal{D}_i} + \Psi_{\mathcal{S}'\mathcal{S}'|\mathcal{D}'_i}\ ,
\end{array}
\end{equation}
which require making a further assumption of conditional independence between $\mathcal{D}'_i$ and $\mathcal{D}_j$ given the support set $\mathcal{S}'$ for $j=1,\ldots,N$.

Finally, to assimilate the local summary of agent $i$ with the other agents' local summaries (i.e., based on their current choices of support sets) into a global summary to be used for predicting the phenomenon, the local summary $(\nu_{\mathcal{S}'|\mathcal{D}_j},\Psi_{\mathcal{S}'\mathcal{S}'|\mathcal{D}_j})$ of
every other agent $j\neq i$ based on agent $i$'s support set $\mathcal{S}'$ can be derived approximately from the received local summary $(\nu_{\mathcal{S}''|\mathcal{D}_j},\Psi_{\mathcal{S}''\mathcal{S}''|\mathcal{D}_j})$ based on agent $j$'s support set $\mathcal{S}''\neq \mathcal{S}'$ using exactly the same transfer learning mechanism described above. Then, the global summary $(\dot{\nu}_{\mathcal{S}'},\dot{\Psi}_{\mathcal{S}'\mathcal{S}'})$ can be computed via~\eqref{eqn:globalsum} and used by the GP-DDF or GP-DDF$^+$ algorithm (Section~\ref{notcareful}).
%
%
\begin{algorithm}[h]\label{alg:local}
\begin{small}
\DontPrintSemicolon
\If{\emph{agent $i$ transits from local area with support set $\mathcal{S}$ to local area with support set $\mathcal{S}'$}} {
\tcc{Transfer learning mechanism}
Construct local summary $(\nu_{\mathcal{S}|\mathcal{D}_i},\Psi_{\mathcal{S}\mathcal{S}|\mathcal{D}_i})$ and transform it to prior summary $(\omega_{\mathcal{S}|\mathcal{D}_i},\Phi_{\mathcal{S}\mathcal{S}|\mathcal{D}_i})$ by~\eqref{eqn:ppl};\;
Derive prior summary $(\omega_{\mathcal{S}'|\mathcal{D}_i},\Phi_{\mathcal{S}'\mathcal{S}'|\mathcal{D}_i})$ based on $\mathcal{S}'$ approximately from $(\omega_{\mathcal{S}|\mathcal{D}_i},\Phi_{\mathcal{S}\mathcal{S}|\mathcal{D}_i})$ by~\eqref{eqn:moving};\;
Transform prior summary $(\omega_{\mathcal{S}'|\mathcal{D}_i},\Phi_{\mathcal{S}'\mathcal{S}'|\mathcal{D}_i})$ to local summary $(\nu_{\mathcal{S}'|\mathcal{D}_i},\Psi_{\mathcal{S}'\mathcal{S}'|\mathcal{D}_i})$ by~\eqref{eqn:ppl};}
\If{\emph{agent $i$ has to predict the phenomenon}} {
\If{\emph{data $(\mathcal{D}'_i, y_{\mathcal{D}'_i})$ is available from local area with support set $\mathcal{S}'$}} {Assimilate local summaries $(\nu_{\mathcal{S}'|\mathcal{D}_i},\Psi_{\mathcal{S}'\mathcal{S}'|\mathcal{D}_i})$ with $(\nu_{\mathcal{S}'|\mathcal{D}'_i},\Psi_{\mathcal{S}'\mathcal{S}'|\mathcal{D}'_i})$ to yield $(\nu_{\mathcal{S}'|\mathcal{D}_i\bigcup\mathcal{D}'_i},\Psi_{\mathcal{S}'\mathcal{S}'|\mathcal{D}_i\bigcup\mathcal{D}'_i})$ by~\eqref{eqn:update};}
Exchange local summary with every agent $j\neq i$;\;
\ForEach{\emph{agent $j\neq i$ in local area with support set $\mathcal{S}''\neq \mathcal{S}'$}}
{Derive local summary $(\nu_{\mathcal{S}'|\mathcal{D}_j},\Psi_{\mathcal{S}'\mathcal{S}'|\mathcal{D}_j})$ based on $\mathcal{S}'$ approximately from received local summary $(\nu_{\mathcal{S}''|\mathcal{D}_j},\Psi_{\mathcal{S}''\mathcal{S}''|\mathcal{D}_j})$ based on $\mathcal{S}''$ using the above transfer learning mechanism;}
Compute global summary $(\dot{\nu}_{\mathcal{S}'},\dot{\Psi}_{\mathcal{S}'\mathcal{S}'})$ by~\eqref{eqn:globalsum} using local summaries $(\nu_{\mathcal{S}'|\mathcal{D}_i\bigcup\mathcal{D}'_i},\Psi_{\mathcal{S}'\mathcal{S}'|\mathcal{D}_i\bigcup\mathcal{D}'_i})$ and $(\nu_{\mathcal{S}'|\mathcal{D}_j},\Psi_{\mathcal{S}'\mathcal{S}'|\mathcal{D}_j})$ of every agent $j\neq i$;\;
Run GP-DDF or GP-DDF$^+$~(Section~\ref{gpm});\;
}
\end{small}
\caption{GP-DDF/GP-DDF$^+$ with agent-centric support sets based on  transfer learning for agent $i$}
\end{algorithm}

Supposing $|\mathcal{S}|=|\mathcal{S}'|=|\mathcal{S}''|$ for simplicity,
our transfer learning mechanism in Algorithm~\ref{alg:local} incurs only $\mathcal{O}(|\mathcal{S}|^3)$ time (i.e., independent of data size $|\mathcal{D}|$) due to multiplication and inversion of matrices of size $|\mathcal{S}|$ by $|\mathcal{S}|$. Since the support set for every local area is expected to be small, our transfer learning mechanism is efficient and scalable.
%
%
\subsubsection{Information Loss from Low-Rank Approximation.}
\label{guarantee}
Recall from the remark after Proposition~\ref{prop0} that our transfer learning mechanism has utilized a low-rank covariance matrix approximation $\Sigma_{\mathcal{S}'\mathcal{S}}\Sigma_{\mathcal{S}\mathcal{S}}^{-1}\Sigma_{\mathcal{S}\mathcal{D}_i}$ of $\Sigma_{\mathcal{S}'\mathcal{D}_i}$. To theoretically bound the information loss resulting from such an approximation, we first observe that it resembles the Nystr{\"{o}}m low-rank approximation except that the latter typically involves approximating a symmetric positive semi-definite matrix like $\Sigma_{\mathcal{S}'\mathcal{S}'}$ or $\Sigma_{\mathcal{D}_i\mathcal{D}_i}$ instead of $\Sigma_{\mathcal{S}'\mathcal{D}_i}$, which precludes a direct application of existing  results on Nystr{\"{o}}m approximation 
to our theoretical analysis.
Fortunately, we can exploit the idea of clustering with respect to $\mathcal{S}$ for our theoretical analysis which is inspired by that of the Nystr{\"{o}}m approximation of \citeauthor{Kwok08}~\shortcite{Kwok08} but results in a different loss bound depending on the GP hyperparameters (Section~\ref{gpm}) and the ``closeness'' of $\mathcal{S}'$ and $\mathcal{D}_i$ to $\mathcal{S}$ in the correlation sense.



Define $c(x)$ as a function mapping each $x\in\mathcal{D}_i\bigcup\mathcal{S}'$ to the ``closest'' $c(x)\in\mathcal{S}$, that is, $c:\mathcal{D}_i\bigcup\mathcal{S}'\rightarrow \mathcal{S}$ where $c(x)\triangleq\mathop{\arg\min}_{s\in\mathcal{S}} ||\Lambda^{-1}(x-s)||$.
Then, partition $\mathcal{D}_i$ ($\mathcal{S}'$) into $|\mathcal{S}|$ disjoint subsets
$\mathcal{D}_{is}\triangleq\{x\in\mathcal{D}_i\mid c(x) =s\}$
($\mathcal{S}'_{s}\triangleq\{x\in \mathcal{S}'\mid c(x) = s\}$)
for $s\in\mathcal{S}$.
Intuitively, $\mathcal{D}_{is}$ ($\mathcal{S}'_{s}$) is a cluster of locations in $\mathcal{D}_i$ ($\mathcal{S}'$) that are closest to location $s$ in the support set $\mathcal{S}$.
%
Our main result below theoretically bounds the information loss $||\Sigma_{\mathcal{S}'\mathcal{D}_i} - \Sigma_{\mathcal{S}'\mathcal{S}}\Sigma_{\mathcal{S}\mathcal{S}}^{-1}\Sigma_{\mathcal{S}\mathcal{D}_i}||_F$ resulting from the low-rank approximation $\Sigma_{\mathcal{S}'\mathcal{S}}\Sigma_{\mathcal{S}\mathcal{S}}^{-1}\Sigma_{\mathcal{S}\mathcal{D}_i}$ of $\Sigma_{\mathcal{S}'\mathcal{D}_i}$
with respect to the Frobenius norm:
\begin{thm}\label{thm1}
Let $\sigma_{xx'}$ be defined by a squared exponential covariance function \eqref{secf1},
$T\triangleq\arg\max_{s\in\mathcal{S}} |\mathcal{D}_{is}|$, $T'\triangleq\arg\max_{s\in\mathcal{S}} |\mathcal{S}'_{s}|$, $\epsilon_{\mathcal{S}'}\triangleq |\mathcal{S}'|^{-1}\sum_{x\in\mathcal{S}'}||\Lambda^{-1}(x -c(x))||^2$, and $\epsilon_{\mathcal{D}_{i}}\triangleq |\mathcal{D}_{i}|^{-1}\sum_{x\in\mathcal{D}_{i}}||\Lambda^{-1}(x-c(x))||^2$. Then, 
$$
\begin{array}{l}
\displaystyle||\Sigma_{\mathcal{S}'\mathcal{D}_i} - \Sigma_{\mathcal{S}'\mathcal{S}}\Sigma_{\mathcal{S}\mathcal{S}}^{-1}\Sigma_{\mathcal{S}\mathcal{D}_i}||_F\leq \sqrt{3/e}\sigma^2_s|\mathcal{S}|TT'(\sqrt{\epsilon_{\mathcal{S}'}}\\
 \displaystyle +\sqrt{\epsilon_{\mathcal{S}'} + \epsilon_{\mathcal{D}_{i}}} +\sqrt{\epsilon_{\mathcal{D}_{i}}}\ + \sigma^2_s||\Sigma_{\mathcal{S}\mathcal{S}}^{-1}||_F|\mathcal{S}|\sqrt{3\epsilon_{\mathcal{S}'}\epsilon_{\mathcal{D}_{i}}/e})\ .
\end{array}
$$
\end{thm}
Its proof is in\if\myproof1 Appendix~\ref{Apx.err}. \fi\if\myproof0 \cite{Ruofei17}. \fi
Note that a similar result to Theorem~\ref{thm1} can be derived for other commonly-used covariance functions such as those presented in the work of \citeauthor{Kwok08}~\shortcite{Kwok08}.
 It can be observed from Theorem~\ref{thm1} that the information loss $||\Sigma_{\mathcal{S}'\mathcal{D}_i} - \Sigma_{\mathcal{S}'\mathcal{S}}\Sigma_{\mathcal{S}\mathcal{S}}^{-1}\Sigma_{\mathcal{S}\mathcal{D}_i}||_F$ can be reduced when the signal variance $\sigma^2_s$ is small, the length-scales $\ell_1$ and/or $\ell_2$ are large,
  the mobile sensing agent $i$ utilizes a support set $\mathcal{S}$ ``close''
to its observed locations $\mathcal{D}_i$ in a local area (i.e., smaller $\epsilon_{\mathcal{D}_{i}}$) and moves to another local area with a support set $\mathcal{S}'$ ``close'' to $\mathcal{S}$ (i.e., smaller $\epsilon_{\mathcal{S}'}$).
%
%
\subsubsection{Lazy Transfer Learning.}
\label{sec:multi}
Theorem~\ref{thm1} above further reveals that every instance of transfer learning in Algorithm~\ref{alg:local} incurs some information loss which accumulates over multiple instances when the agent transits between many local areas and consequently degrades its resulting predictive performance. This motivates the need to be frugal in the number of instances of transfer learning to be performed.

To achieve this, our key idea is to delay transfer learning till prediction time but in a memory-efficient manner\footnote{Naively, an agent can delay transfer learning by simply storing a separate local summary based on the support set for every previously visited local area, which is not memory-efficient.}.
%
Specifically,
we propose the following new information sharing mechanism to reduce memory requirements for a team of mobile sensing agents:
When agent $i$ leaves a local area, its local summary is communicated to another agent in the same area who assimilates it with its own local summary using~\eqref{eqn:globalsum}.
However, if no other agent is in the same area, then agent $i$ stores a backup of its local summary.
%
On the other hand, when agent $i$ enters a local area containing other agents, it simply obtains its corresponding support set to encapsulate its new data gathered in this area.
But, if no other agent is in this area, then agent $i$ retrieves (and removes) the backup of its corresponding local summary from an agent who has previously visited this area\footnote{Multiple backups of the local summary for the same local area may exist if agents leave this area at the same time, which rarely happens. In this case, agent $i$ should retrieve (and remove) all these backups from the agents storing them.}.
If no agent has such a backup, then agent $i$ is the first to visit this area and constructs a new support set for it.
Algorithm\if\myproof1~\ref{alg:sendrecv} (Appendix~\ref{runforest}) \fi\if\myproof0 $2$ in \cite{Ruofei17} \fi
details GP-DDF/GP-DDF$^+$ with agent-centric support sets by incorporating the above information sharing mechanism in order to achieve memory-efficient lazy transfer learning.

To analyze the memory requirements of our information sharing mechanism in Algorithm\if\myproof1~\ref{alg:sendrecv} (Appendix~\ref{runforest}), \fi\if\myproof0 $2$ in \cite{Ruofei17}, \fi
let the domain of the phenomenon be partitioned into $K$ local areas. Then,
the team of $N$ mobile sensing agents incurs a total of $\mathcal{O}((K+N)|\mathcal{S}|^2)$ memory in the worst case when all the agents reside in the same local area and the last agent entering this area stores the backups of the local summaries for the other $K-1$ local areas.
However,
the agents are usually well-distributed over the entire phenomenon in practice:
In the case of evenly distributed agents, the team incurs a total of
$\mathcal{O}(\max(K,N)|\mathcal{S}|^2)$ memory.
So, each agent incurs an amortized memory cost of $\mathcal{O}(\max(K,N)|\mathcal{S}|^2/N)$.

A limitation of the information sharing mechanism in Algorithm\if\myproof1~\ref{alg:sendrecv} (Appendix~\ref{runforest}) \fi\if\myproof0 $2$ in \cite{Ruofei17} \fi
is its susceptibility to agent failure: If an agent stores the backups of the local summaries for many local areas and breaks down, then all the information on these local areas will be lost.
Its robustness to agent failure can be improved by distributing multiple agents to every local area to reduce its risk of being empty and hence its likelihood of inducing a backup.
\section{Experiments and Discussion}
\label{sect:expt}
This section empirically evaluates the performance of our GP-DDF and GP-DDF$^+$ algorithms with agent-centric support sets using simulated spatial phenomena (Section~\ref{toy}) and two real-world environmental phenomena (Section~\ref{realworld}).
\subsubsection{Performance Metrics.} Two performance metrics are used in our experiments: (a) \emph{Root-mean-square error} (RMSE) $\sqrt{|\mathcal{X}|^{-1}\sum_{x \in\mathcal{X}} (\overline{\mu}_x - y_x)^2}$ measures the predictive performance of the tested algorithms while (b) incurred time measures their efficiency and scalability.
\subsection{Simulated Spatial Phenomena}
\label{toy}
%
The simulated experiment here is set up to demonstrate the effectiveness of our proposed lazy transfer learning mechanism (Section~\ref{sec:multi}) that is driving our GP-DDF/GP-DDF$^+$ algorithms with agent-centric support sets\if\myproof1 (Appendix~\ref{runforest}): \fi\if\myproof0 \cite{Ruofei17}: \fi
A number of $2$-dimensional spatial phenomena of size $50$ by $50$ are generated  using signal variance $\sigma^2_s =1$, noise variance $\sigma^2_n = 0.01$, and by varying the length-scale $\ell_1 = \ell_2$ from $1$ to $20$.
The domain of the spatial phenomenon is partitioned into $4$ disjoint local areas of size $25$ by $25$ (Fig.~\ref{fig:demo}), each of which contains an agent moving randomly within  to gather $25$ local data/observations.
We compare the predictive performance of the following decentralized data fusion algorithms:
(a) Original GP-DDF \cite{LowUAI12,LowTASE15} with a common support set of size $18$ uniformly distributed over the entire phenomenon and known to all $4$ agents,
(b) \emph{PITCs utilizing local information} (local PITCs) with agent-centric support sets assign a different PITC to each agent summarizing its gathered local data based on a support set of size $18$ uniformly distributed over its residing local area,
(c) \emph{PITCs utilizing full information} (full PITCs) with agent-centric support sets assign a different PITC to each agent summarizing its gathered local data as well as those communicated by the other agents (i.e., full data gathered by all agents)
based on a support set of size $18$ uniformly distributed over its residing local area,
(d) \emph{GP-DDF with agent-centric support sets} (GP-DDF-ASS) each of size $18$ and uniformly distributed\footnote{\label{boo}Alternatively, active learning can be used to select an informative support set \emph{a priori} for each local area~\cite{LowTASE15}. Empirically, this yields little performance improvement due to a sufficiently dense (yet small) support set uniformly distributed over the local area and slightly beyond its boundary by $10\%$ of its width.}
over a different local area (Algorithm\if\myproof1~\ref{alg:sendrecv} in Appendix~\ref{runforest}). \fi\if\myproof0 $2$ in \cite{Ruofei17}). \fi
Note that if our proposed lazy transfer learning mechanism in GP-DDF-ASS incurs minimal (total) information loss, then
its predictive performance will be similar to that of full PITCs (local PITCs).
%
%
%

Fig.~\ref{fig:demo} shows results of the maps of log-predictive variance (i.e., $\log\overline{\sigma}^2_{x}$ for all $x\in\mathcal{X}$) over a spatial phenomenon with length-scale of $10$ achieved by the tested decentralized data fusion algorithms.
It can be observed from Fig.~\ref{fig:demo}a that GP-DDF achieves the worst predictive performance since its common support set, which is uniformly distributed over the entire phenomenon, is of the same size as an agent-centric support set uniformly distributed over each of the $4$ smaller disjoint local areas to be used by the other tested algorithms.
From Fig.~\ref{fig:demo}b, though local PITCs can predict better than GP-DDF, the predictive uncertainty at the boundaries between local areas remains very high, which is previously explained in Section~\ref{sect:tl}.
Fig.~\ref{fig:demo}c shows the most ideal predictive performance achieved by full PITCs  because each agent exploits the full data gathered by and exchanged with all agents for encapsulating into a global summary based on the support set distributed over its residing local area.
Fig.~\ref{fig:demo}d reveals that GP-DDF-ASS can achieve predictive performance comparable to that of full PITCs without needing to exchange the full data between all agents due to minimal information loss by our lazy transfer learning mechanism.

%

Recall from Theorem~\ref{thm1} (Section~\ref{guarantee}) that the information loss incurred by our proposed transfer learning mechanism depends on the closeness between the support sets distributed over different local areas as well as the closeness (i.e., in the correlation sense) between the support sets and the data/observations.
The effect of varying such closeness on the performance of our  transfer learning mechanism can be empirically investigated by alternatively changing the length-scale to control the degree of spatial correlation between the measurements of the phenomenon.
Fig.~\ref{fig:demo}e shows results of the reduction in RMSE of GP-DDF, full PITCs, and GP-DDF-ASS over local PITCs with varying lengthscales from $1$ to $20$.
It can be observed that only GP-DDF performs worse than local PITCs while both GP-DDF-ASS and full PITCs perform significantly better than local PITCs, all of which are explained previously.
Interestingly, the reduction in RMSEs varies for different length-scales
and tends to zero when the length-scale is either too small or large.
With a very small length-scale, the correlations between the support sets distributed over different local areas and between the support sets and the data/observations become near-zero, hence resulting in poor transfer learning for GP-DDF-ASS. This agrees with the observation in our theoretical analysis for Theorem~\ref{thm1} (Section~\ref{guarantee}).
With a very large length-scale, though their correlations are strong, the local observations/data can be used by local PITCs to predict very well, hence making transfer learning redundant.
Our transfer learning mechanism performs best with intermediate length-scales where the correlations between the support sets distributed over different local areas and between the support sets and the data are sufficiently strong but not to the extent of achieving good predictions with simply local data.
%
\subsection{Real-World Environmental Phenomena}
\label{realworld}
The performance of our GP-DDF and GP-DDF$^+$ algorithms with agent-centric support sets are empirically evaluated using the following two real-world datasets (as well as the MODIS plankton density dataset in\if\myproof1 Appendix~\ref{plankton}): \fi\if\myproof0 \cite{Ruofei17}): \fi
%
%
(a) The indoor lighting quality dataset contains $1200$ observations of relative lighting level gathered simultaneously by three real Pioneer $3$-DX mobile robots mounted with SICK LMS$200$ laser rangefinders and weather boards while patrolling an office environment, as shown in\if\myproof1 Appendix~\ref{robot}. \fi\if\myproof0 \cite{Ruofei17}. \fi
The domain of interest is partitioned into $K=8$ consecutive local areas
and the robots patrol to and fro across them such that they visit all $K=8$ local areas exactly twice to gather observations of relative lighting level; and
(b) the monthly sea surface temperature ($^{\circ}$C) dataset\if\myproof1 (Appendix~\ref{temperature}) \fi\if\myproof0 \cite{Ruofei17} \fi
is bounded within lat. $35.75$-$14.25$S and lon. $80.25$-$104.25$E (i.e., in the Indian ocean) and gathered from Dec. $2002$ to Dec. $2015$ with a data size of $1083608$.
The huge spatiotemporal domain of this phenomenon comprises
$5$-dimensional input feature vectors of latitude, longitude, year, month, and season,
and is spatially partitioned into $32$ disjoint local areas, each of which is temporally split into $64$ disjoint intervals (hence, $K=2048$) and assigned $2$ agents moving randomly within to gather local observations (hence, a total of $64$ agents); the results are averaged over $10$ runs.
\begin{figure*}
\centering
\begin{tabular}{ccccc}
  \includegraphics[scale=0.175]{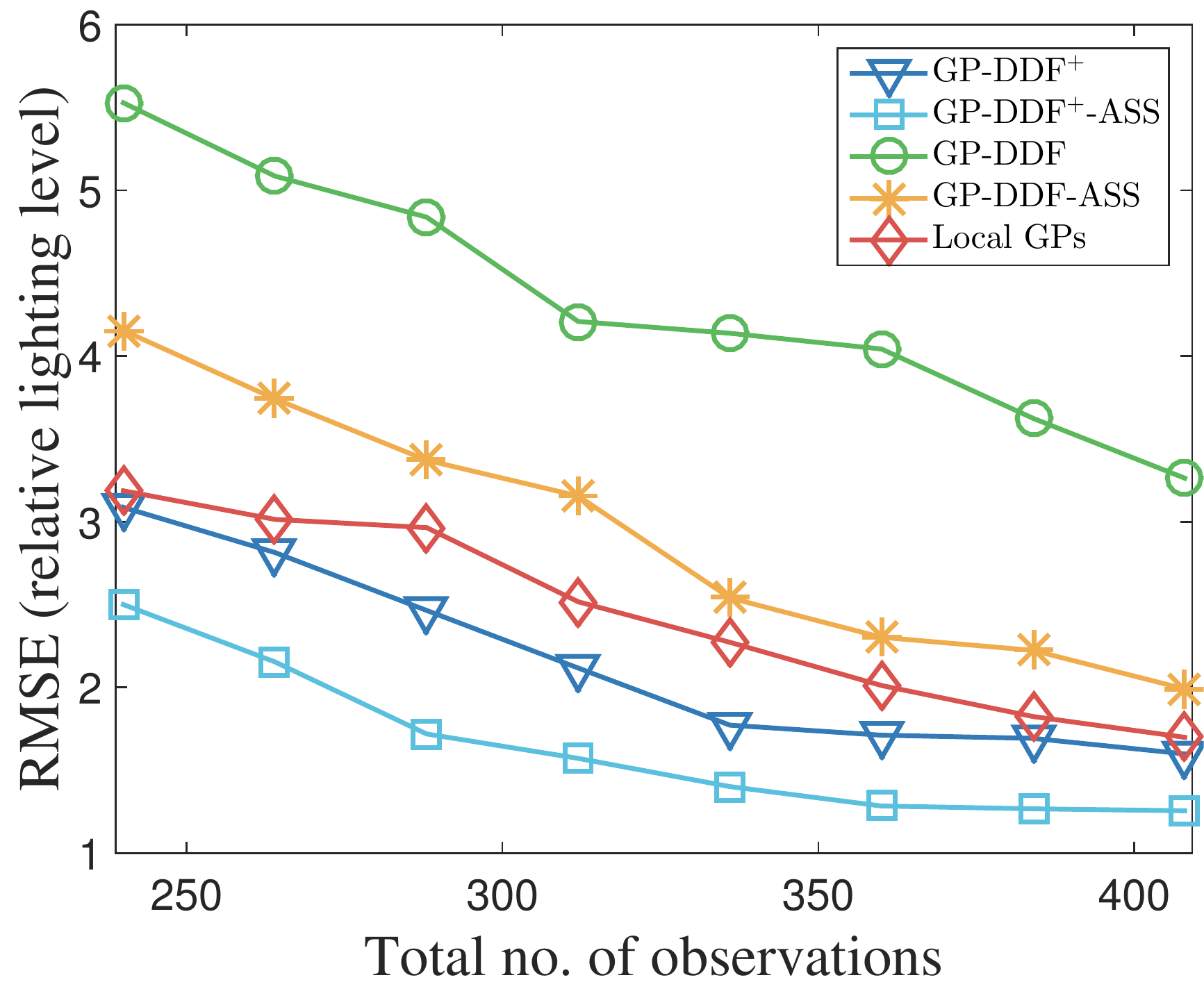} & 
  \includegraphics[scale=0.175]{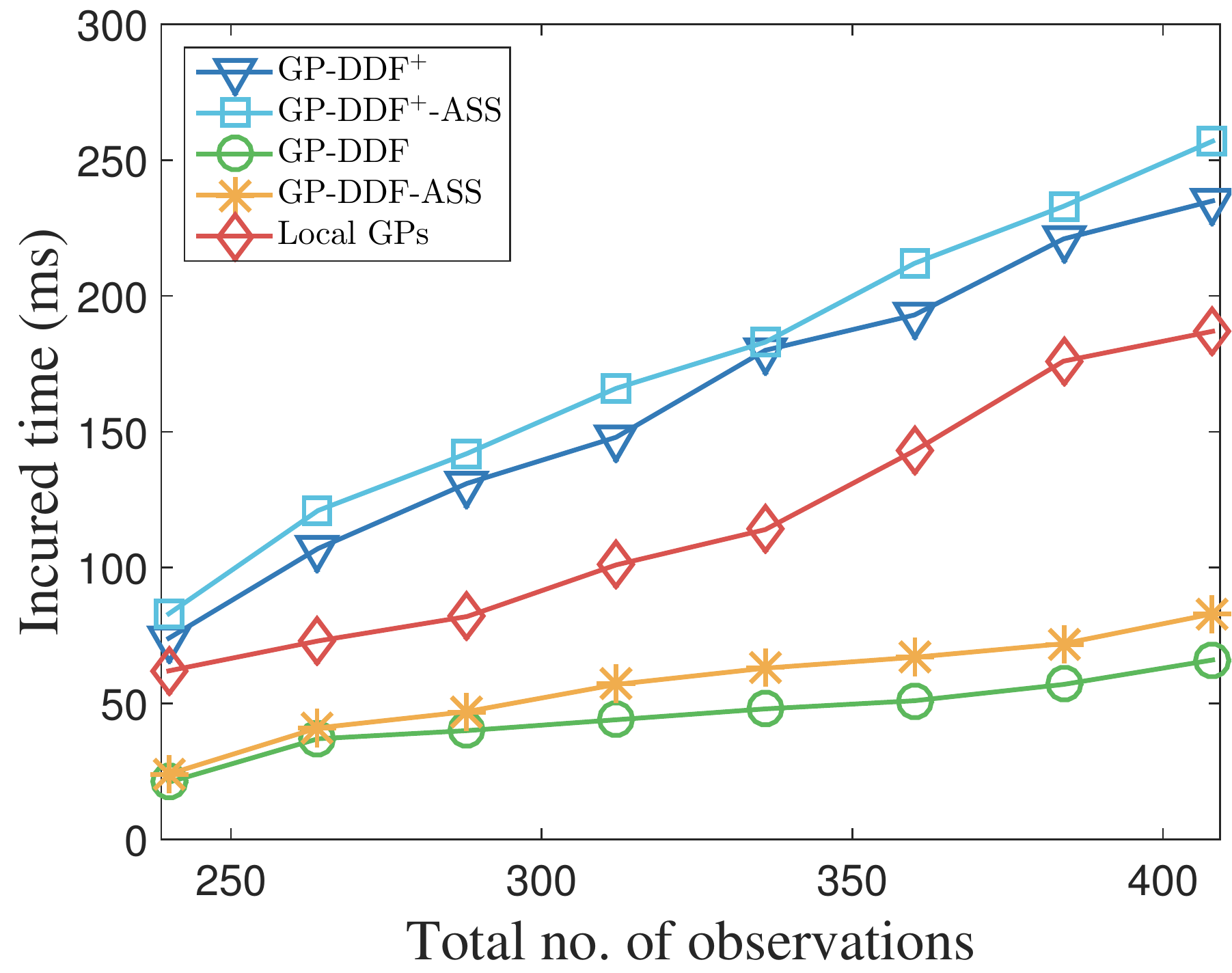} &
  \includegraphics[scale=0.175]{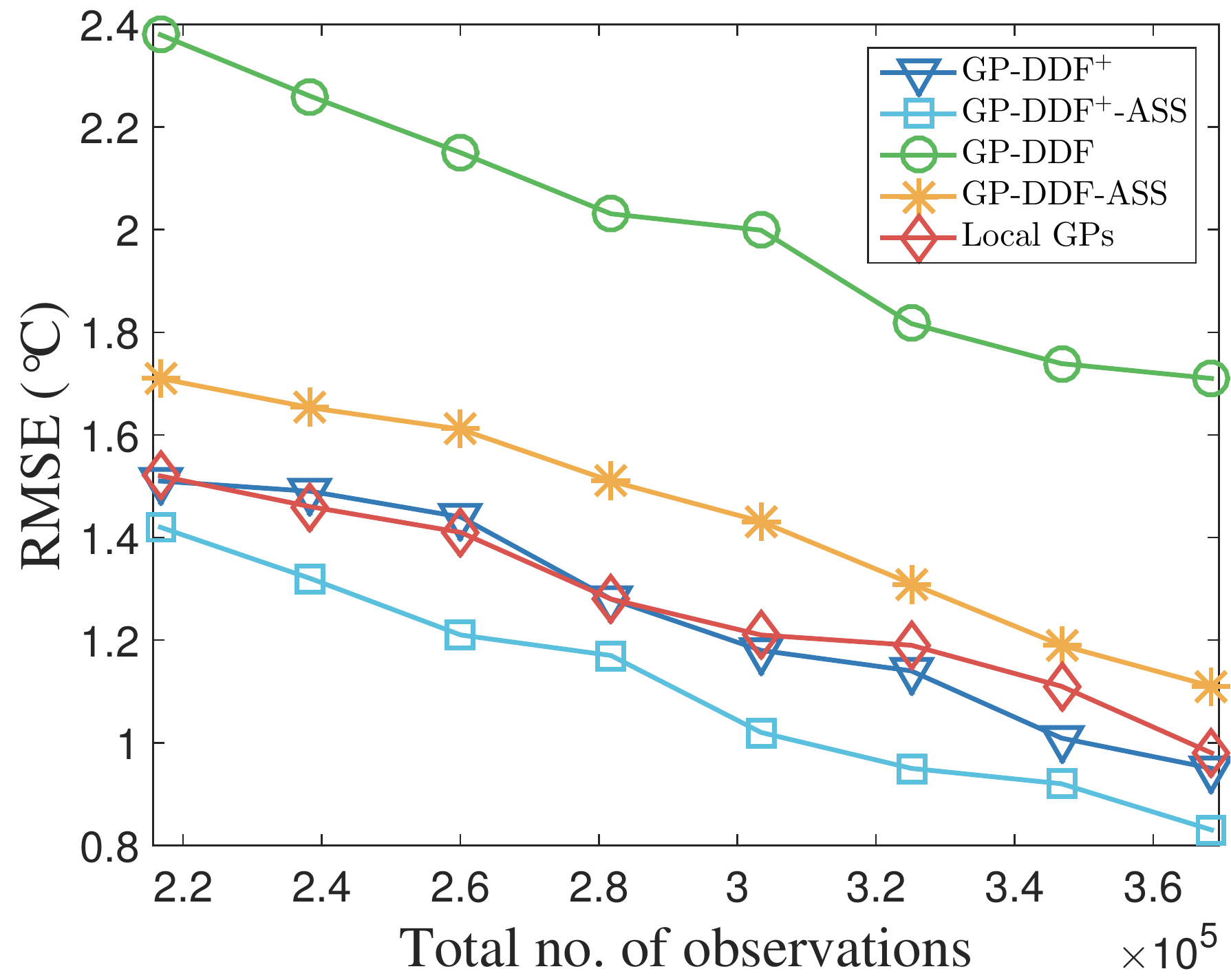} &
  \includegraphics[scale=0.175]{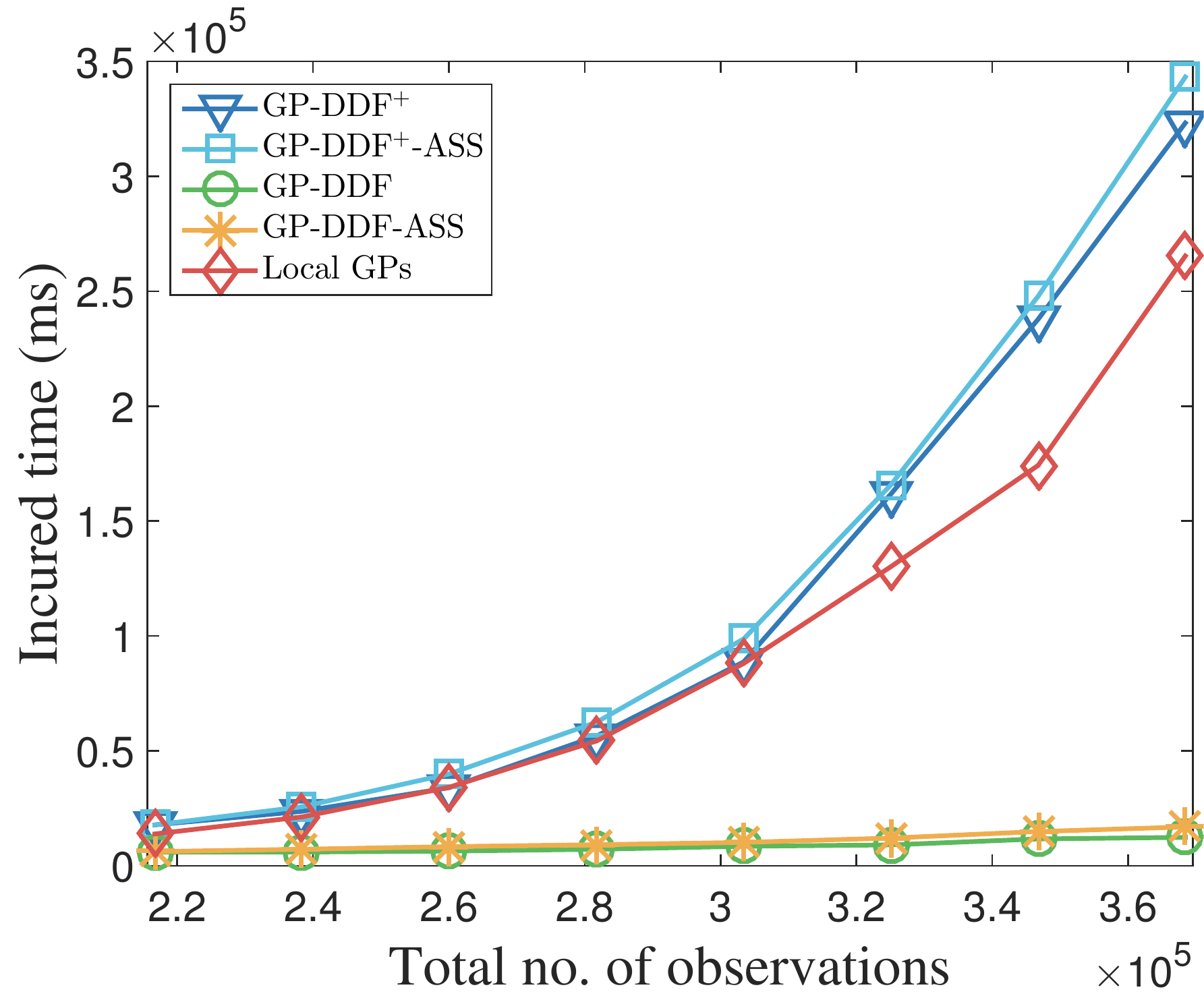} &
  \includegraphics[scale=0.175]{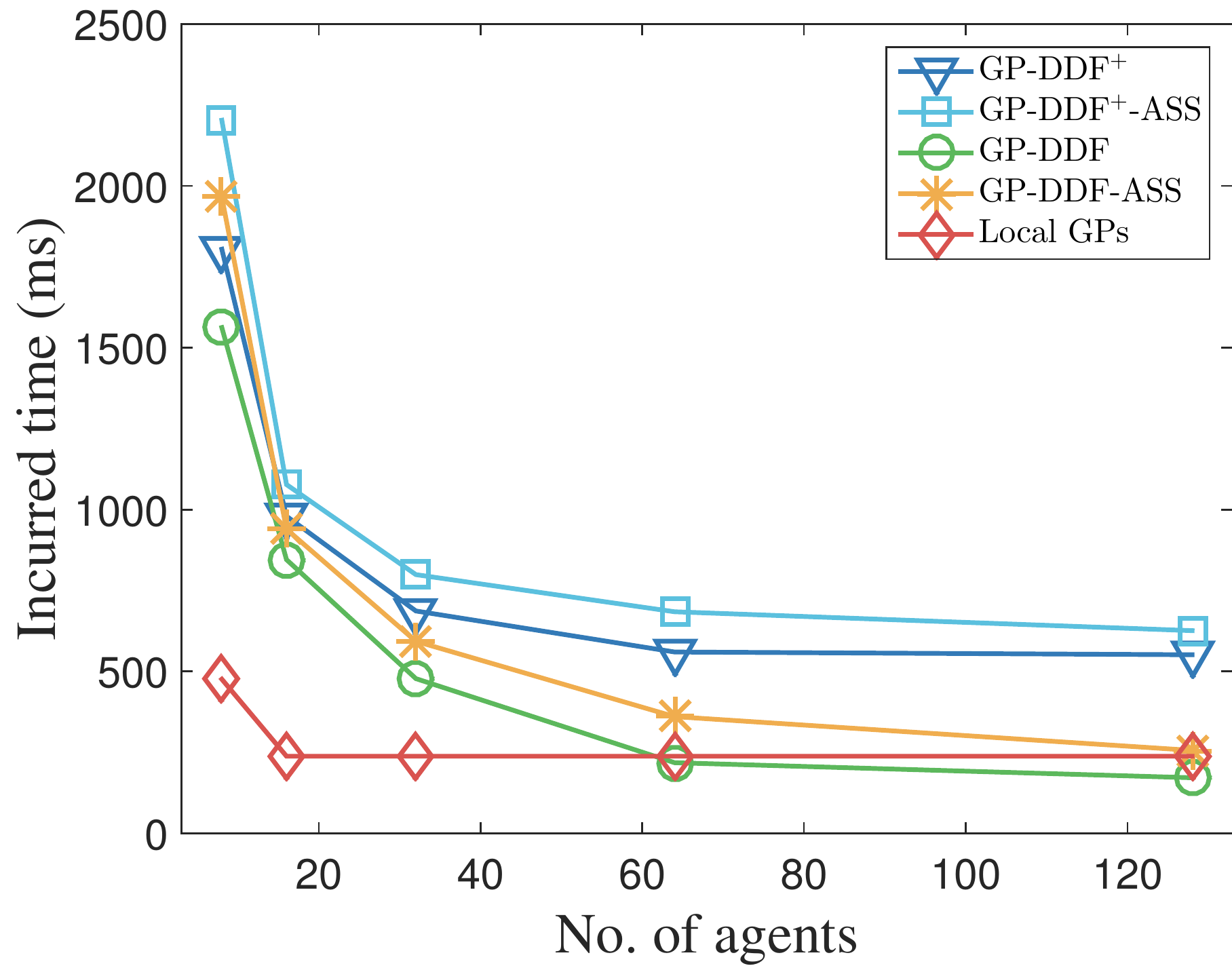}\\
  (a)  &  (b) &  (c) & (d) & (e) 
\end{tabular}
\caption{Graphs of RMSE and total time incurred by tested algorithms vs. total no. of observations for (a-b) indoor lighting quality and (c-d) temperature phenomenon, and (e) graphs of total incurred time vs. no. of agents achieved by tested algorithms for plankton density phenomenon.}
\label{fig:rmse}
\end{figure*}



%
The performance of our \emph{GP-DDF and GP-DDF$^+$ algorithms with agent-centric support sets} (respectively, GP-DDF-ASS and GP-DDF$^+$-ASS),
each of which is of size $64$ ($324$) and uniformly distributed\cref{boo} over a different local area of the office environment (temperature phenomenon),
are compared against that of the local GPs\footnote{Local GPs result from a sparse block-diagonal $\Sigma_{\mathcal{D}\mathcal{D}}$~\eqref{socareless}.} method 
and state-of-the-art GP-DDF and GP-DDF$^+$ \cite{LowTASE15} with a common support set of size $64$ ($324$) uniformly distributed over the entire office environment (temperature phenomenon) and known to all agents; consequently, the latter construct local summaries of the same size. The hyperparameters of GP-DDF-ASS and GP-DDF$^+$-ASS are learned using maximum likelihood estimation, as detailed in\if\myproof1 Appendix~\ref{hyper}. \fi\if\myproof0 \cite{Ruofei17}. \fi


\subsubsection{Predictive Performance.}
%
Figs.~\ref{fig:rmse}a and~\ref{fig:rmse}c show results of decreasing RMSE achieved by tested algorithms with an increasing total number of observations, which is expected.
It can be observed that GP-DDF-ASS and GP-DDF$^+$-ASS, respectively, outperform GP-DDF and GP-DDF$^+$, as explained previously in the last paragraph of Section~\ref{sec:intro}.
Furthermore, the performance improvement of GP-DDF-ASS over GP-DDF is larger than that of GP-DDF$^+$-ASS over GP-DDF$^+$, which demonstrates the effectiveness of our lazy transfer learning mechanism, especially when
some local areas lack data/observations.
This also explains the better predictive performance of GP-DDF$^+$-ASS over local GPs, even though they both exploit local data.

\subsubsection{Time Efficiency.} In this experiment, we specifically evaluate the time efficiency of our transfer learning mechanism (Section~\ref{sec:multi}) in GP-DDF-ASS and GP-DDF$^+$-ASS with respect to the number of observations; to do this, we have intentionally ignored the time incurred by their information sharing mechanism (i.e., first if-then construct in Algorithm 
\if\myproof1~\ref{alg:sendrecv} in Appendix~\ref{runforest}) \fi\if\myproof0 $2$ in \cite{Ruofei17}) \fi
and compared their resulting incurred time with that of GP-DDF and GP-DDF$^+$ (i.e., without transfer learning).
Figs.~\ref{fig:rmse}b and~\ref{fig:rmse}d show results of increasing total time incurred by tested algorithms when the total number of observations increases, which is expected (Section~\ref{notcareful}).
It can be observed that GP-DDF-ASS and GP-DDF$^+$-ASS, respectively,  incur only slightly more time than GP-DDF and GP-DDF$^+$ (i.e., due to an extra small fixed cost of $\mathcal{O}(|\mathcal{S}|^3)$ time for transfer learning (Section~\ref{gpddfass})) to achieve more superior predictive performance, especially for GP-DDF-ASS.
GP-DDF$^+$-ASS incurs more time than GP-DDF-ASS (local GPs) to further exploit local data (support set and transfer learning) for improving its predictive performance.
For \emph{time-critical applications}, we recommend using GP-DDF-ASS over GP-DDF$^+$-ASS since its incurred time is small and increases very gradually with more observations while its performance improvement over GP-DDF is significant.
For \emph{big data applications}, GP-DDF$^+$-ASS is instead preferred since a large amount of local data is often available in nearly every local area for prediction.

\subsubsection{Scalability in the Number of Agents.} Fig.~\ref{fig:rmse}e shows results of total time incurred by tested algorithms averaged over $30$ runs with an increasing number $N$ of agents (i.e., up to $128$ agents) to gather a total number of $1235$ observations from a plankton density phenomenon; the experimental setup is detailed in\if\myproof1 Appendix~\ref{plankton}. \fi\if\myproof0 \cite{Ruofei17}. \fi
It can be observed that the total time incurred by GP-DDF-ASS and GP-DDF$^+$-ASS decrease with more agents, as explained in Section~\ref{notcareful}, and they, respectively,  incur only slightly more time than GP-DDF and GP-DDF$^+$ due to their information sharing mechanism described in Section~\ref{gpddfass} (i.e., first if-then construct in Algorithm 
\if\myproof1~\ref{alg:sendrecv} in Appendix~\ref{runforest}). \fi\if\myproof0 $2$ in \cite{Ruofei17}). \fi
Additional empirical results and analysis for the plankton density phenomenon are reported in\if\myproof1 Appendix~\ref{plankton}. \fi\if\myproof0 \cite{Ruofei17}. \fi
%
\section{Conclusion}
This paper describes novel GP-DDF-ASS and GP-DDF$^+$-ASS algorithms for distributed cooperative perception of large-scale environmental phenomena. To overcome the limitations of scale of GP-DDF and GP-DDF$^+$,
%
our proposed algorithms employ a novel transfer learning mechanism between agents which is capable of sharing and  transferring  information encapsulated in a summary based on a support set to that utilizing a different support set with some loss that can be theoretically bounded and analyzed.
To alleviate the issue of information loss accumulating over multiple instances of transfer learning, GP-DDF-ASS and GP-DDF$^+$-ASS exploit a new information sharing mechanism to achieve memory-efficient lazy transfer learning.
Empirical evaluation on real-world datasets show that our transfer learning and information sharing mechanisms make GP-DDF-ASS and GP-DDF$^+$-ASS incur only slightly more time than GP-DDF and GP-DDF$^+$ (i.e., without transfer learning) to achieve more superior predictive performance.
%

\subsubsection{Acknowledgments.}
This research is supported by Singapore Ministry of Education Academic Research Fund Tier~$2$, MOE$2016$-T$2$-$2$-$156$.

\bibliographystyle{aaai}
\bibliography{dydess2014}

\if \myproof1
\clearpage
\appendix
\section{Gaussian Predictive Distribution computed by the GP-DDF$^+$ Algorithm}
\label{great}
\begin{defn}[GP-DDF$^+$]\label{def:gpddfp}
Given a common support set $\mathcal{S}\subset\mathcal{X}$ known to all $N$ agents,  global summary $(\dot{\nu}_{\mathcal{S}},\dot{\Psi}_{\mathcal{S}\mathcal{S}})$, local summary $(\nu_{\mathcal{S}|\mathcal{D}_i},\Psi_{\mathcal{S}\mathcal{S}|\mathcal{D}_i})$, and a column vector $y_{\mathcal{D}_i}$ of realized measurements for observed locations $\mathcal{D}_i$, the GP-DDF$^+$ algorithm run by each agent $i$ computes a  Gaussian predictive distribution $\mathcal{N}(\overline{\mu}_{x}, \overline{\sigma}^2_{x})$ of the measurement for any unobserved location $x \in \mathcal{X} \setminus \mathcal{D}$ where
\begin{equation}\label{supercareless}\hspace{-1mm}
\begin{array}{rcl}
\overline{\mu}_{x} \hspace{-2.4mm}&\triangleq & \hspace{-2.4mm}\displaystyle \mu_{x}+\left(\gamma_{x\mathcal{S}}^i\dot{\Psi}^{-1}_{\mathcal{S}\mathcal{S}}\dot{\nu}_{\mathcal{S}} -\Sigma_{x\mathcal{S}}\Sigma_{\mathcal{S}\mathcal{S}}^{-1}\nu_{\mathcal{S}|\mathcal{D}_i}\right)+{\nu}_{x|\mathcal{D}_i}\ ,\vspace{0.5mm}\\
\overline{\sigma}^2_{x} \hspace{-2.4mm}&\triangleq & \hspace{-2.4mm}\displaystyle  \sigma_{xx} - \Big(\gamma_{x\mathcal{S}}^i\Sigma_{\mathcal{S}\mathcal{S}}^{-1}\Sigma_{\mathcal{S}x}-\Sigma_{x\mathcal{S}}\Sigma_{\mathcal{S}\mathcal{S}}^{-1}\Psi_{\mathcal{S}x|\mathcal{D}_i}\vspace{0.5mm}\\
&& \hspace{-2.4mm}\displaystyle\ \quad\qquad -\ \gamma_{x\mathcal{S}}^i\dot{\Psi}_{\mathcal{S}\mathcal{S}}^{-1}\gamma_{\mathcal{S}x}^i \Big)-\Psi_{xx|\mathcal{D}_i}\ ,
\end{array}\hspace{-4mm}
\end{equation}
$\gamma_{x\mathcal{S}}^i \triangleq \displaystyle \Sigma_{x\mathcal{S}}+\Sigma_{x\mathcal{S}}\Sigma_{\mathcal{S}\mathcal{S}}^{-1}\Psi_{\mathcal{S}\mathcal{S}|\mathcal{D}_i}-\Psi_{x\mathcal{S}|\mathcal{D}_i}\ ,$ and $\gamma_{\mathcal{S}x}^i \triangleq\gamma_{x\mathcal{S}}^{i\top}\ .$
\end{defn}
The Gaussian predictive distribution~\eqref{supercareless} computed by the GP-DDF$^+$ algorithm is observed to exploit the local and global summaries (i.e., terms within brackets) as well as the data local to agent $i$ (i.e., ${\nu}_{x|\mathcal{D}_i}$ and $\Psi_{xx|\mathcal{D}_i}$ terms).
\section{Proof of Proposition~\ref{prop0}}
\label{Apx.ppt}
%
\begin{equation*}
\begin{array}{l}
\omega_{\mathcal{S}'|\mathcal{D}_i}\\
=\displaystyle\Sigma_{\mathcal{S}'\mathcal{D}_i}\Sigma_{\mathcal{D}_i\mathcal{D}_i}^{-1}(y_{\mathcal{D}_i}-\mu_{\mathcal{D}_i})\\
= \displaystyle\Sigma_{\mathcal{S}'\mathcal{S}}\Sigma_{\mathcal{S}\mathcal{S}}^{-1}\Sigma_{\mathcal{S}\mathcal{D}_i}\Sigma_{\mathcal{D}_i\mathcal{D}_i}^{-1}(y_{\mathcal{D}_i}-\mu_{\mathcal{D}_i})\\
= \displaystyle\Sigma_{\mathcal{S}'\mathcal{S}}\Sigma_{\mathcal{S}\mathcal{S}}^{-1}\omega_{\mathcal{S}|\mathcal{D}_i}
\end{array}
\end{equation*}
and
\begin{equation*}
\begin{array}{l}
\Phi_{\mathcal{S}'\mathcal{S}'|\mathcal{D}_i}\\
=\displaystyle\Sigma_{\mathcal{S}'\mathcal{D}_i}\Sigma_{\mathcal{D}_i\mathcal{D}_i}^{-1}\Sigma_{\mathcal{D}_i\mathcal{S}'}\\
= \displaystyle\Sigma_{\mathcal{S}'\mathcal{S}}\Sigma_{\mathcal{S}\mathcal{S}}^{-1}\Sigma_{\mathcal{S}\mathcal{D}_i}\Sigma_{\mathcal{D}_i\mathcal{D}_i}^{-1}\Sigma_{\mathcal{D}_i\mathcal{S}}\Sigma_{\mathcal{S}\mathcal{S}}^{-1}\Sigma_{\mathcal{S}\mathcal{S}'}\\
= \displaystyle\Sigma_{\mathcal{S}'\mathcal{S}}\Sigma_{\mathcal{S}\mathcal{S}}^{-1}\Phi_{\mathcal{S}\mathcal{S}|\mathcal{D}_i}\Sigma_{\mathcal{S}\mathcal{S}}^{-1}\Sigma_{\mathcal{S}\mathcal{S}'}
\end{array}
\end{equation*}
where the second equalities above follow from the assumption that
$\mathcal{S}'$ and $\mathcal{D}_i$ are conditionally independent given $\mathcal{S}$ (i.e., $\Sigma_{\mathcal{S}'\mathcal{D}_i|\mathcal{S}}=  \Sigma_{\mathcal{S}'\mathcal{D}_i} - \Sigma_{\mathcal{S}'\mathcal{S}}\Sigma_{\mathcal{S}\mathcal{S}}^{-1}\Sigma_{\mathcal{S}\mathcal{D}_i} =\underline{0}$).
\section{Proof of Proposition~\ref{prop1}}\label{Apx.l2p}
\begin{equation*}
\hspace{-1.7mm}
\begin{array}{l}
\Psi_{\mathcal{S}\mathcal{S}|\mathcal{D}_i}\\
=\displaystyle\Sigma_{\mathcal{S}\mathcal{D}_i}\Sigma_{\mathcal{D}_i\mathcal{D}_i|\mathcal{S}}^{-1}\Sigma_{\mathcal{D}_i\mathcal{S}}\\
=\displaystyle\Sigma_{\mathcal{S}\mathcal{D}_i}(\Sigma_{\mathcal{D}_i\mathcal{D}_i}^{-1}+\Sigma_{\mathcal{D}_i\mathcal{D}_i}^{-1}\Sigma_{\mathcal{D}_i\mathcal{S}}\Sigma_{\mathcal{S}\mathcal{S}|\mathcal{D}_i}^{-1}\Sigma_{\mathcal{S}\mathcal{D}_i}\Sigma_{\mathcal{D}_i\mathcal{D}_i}^{-1})\Sigma_{\mathcal{D}_i\mathcal{S}}\\
=\displaystyle\Sigma_{\mathcal{S}\mathcal{D}_i}\Sigma_{\mathcal{D}_i\mathcal{D}_i}^{-1}\Sigma_{\mathcal{D}_i\mathcal{S}}\ +\vspace{0.5mm}\\
\displaystyle\quad\Sigma_{\mathcal{S}\mathcal{D}_i}\Sigma_{\mathcal{D}_i\mathcal{D}_i}^{-1}\Sigma_{\mathcal{D}_i\mathcal{S}}\Sigma_{\mathcal{S}\mathcal{S}|\mathcal{D}_i}^{-1}\Sigma_{\mathcal{S}\mathcal{D}_i}\Sigma_{\mathcal{D}_i\mathcal{D}_i}^{-1}\Sigma_{\mathcal{D}_i\mathcal{S}}\\
=\displaystyle\Phi_{\mathcal{S}\mathcal{S}|\mathcal{D}_i}+\Phi_{\mathcal{S}\mathcal{S}|\mathcal{D}_i}(\Sigma_{\mathcal{S}\mathcal{S}}-\Sigma_{\mathcal{S}\mathcal{D}_i}\Sigma_{\mathcal{D}_i\mathcal{D}_i}^{-1}\Sigma_{\mathcal{D}_i\mathcal{S}})^{-1}\Phi_{\mathcal{S}\mathcal{S}|\mathcal{D}_i}\\
=\displaystyle\Phi_{\mathcal{S}\mathcal{S}|\mathcal{D}_i}+\Phi_{\mathcal{S}\mathcal{S}|\mathcal{D}_i}(\Sigma_{\mathcal{S}\mathcal{S}}-\Phi_{\mathcal{S}\mathcal{S}|\mathcal{D}_i})^{-1}\Phi_{\mathcal{S}\mathcal{S}|\mathcal{D}_i}\\
=\displaystyle\Phi_{\mathcal{S}\mathcal{S}|\mathcal{D}_i}(I+(\Sigma_{\mathcal{S}\mathcal{S}}-\Phi_{\mathcal{S}\mathcal{S}|\mathcal{D}_i})^{-1}\Phi_{\mathcal{S}\mathcal{S}|\mathcal{D}_i})\\
=\displaystyle\Phi_{\mathcal{S}\mathcal{S}|\mathcal{D}_i}(\Sigma_{\mathcal{S}\mathcal{S}}-\Phi_{\mathcal{S}\mathcal{S}|\mathcal{D}_i})^{-1}(\Sigma_{\mathcal{S}\mathcal{S}}-\Phi_{\mathcal{S}\mathcal{S}|\mathcal{D}_i}+\Phi_{\mathcal{S}\mathcal{S}|\mathcal{D}_i})\\
=\displaystyle\Phi_{\mathcal{S}\mathcal{S}|\mathcal{D}_i}(\Sigma_{\mathcal{S}\mathcal{S}}-\Phi_{\mathcal{S}\mathcal{S}|\mathcal{D}_i})^{-1}\Sigma_{\mathcal{S}\mathcal{S}}
\end{array}
\end{equation*}
where the second equality follows from the matrix inverse lemma on $\Sigma_{\mathcal{D}_i\mathcal{D}_i|\mathcal{S}}^{-1} = (\Sigma_{\mathcal{D}_i\mathcal{D}_i}-\Sigma_{\mathcal{D}_i\mathcal{S}}\Sigma_{\mathcal{S}\mathcal{S}}^{-1}\Sigma_{\mathcal{S}\mathcal{D}_i})^{-1} =\Sigma_{\mathcal{D}_i\mathcal{D}_i}^{-1}+\Sigma_{\mathcal{D}_i\mathcal{D}_i}^{-1}\Sigma_{\mathcal{D}_i\mathcal{S}}\Sigma_{\mathcal{S}\mathcal{S}|\mathcal{D}_i}^{-1}\Sigma_{\mathcal{S}\mathcal{D}_i}\Sigma_{\mathcal{D}_i\mathcal{D}_i}^{-1}$.
As a result, $\Psi_{SS|\mathcal{D}_i}^{-1}=\Sigma_{SS}^{-1}(\Sigma_{SS}-\Phi_{SS|\mathcal{D}_i})\Phi_{SS|\mathcal{D}_i}^{-1} =\Phi_{SS|\mathcal{D}_i}^{-1} - \Sigma_{SS}^{-1}$.
\begin{equation*}
\hspace{-1.7mm}
\begin{array}{l}
\nu_{\mathcal{S}|\mathcal{D}_i}\\
=\displaystyle\Sigma_{\mathcal{S}\mathcal{D}_i}\Sigma_{\mathcal{D}_i\mathcal{D}_i|\mathcal{S}}^{-1}y_{\mathcal{D}_i}\\
=\displaystyle\Sigma_{\mathcal{S}\mathcal{D}_i}(\Sigma_{\mathcal{D}_i\mathcal{D}_i}^{-1}+\Sigma_{\mathcal{D}_i\mathcal{D}_i}^{-1}\Sigma_{\mathcal{D}_i\mathcal{S}}\Sigma_{\mathcal{S}\mathcal{S}|\mathcal{D}_i}^{-1}\Sigma_{\mathcal{S}\mathcal{D}_i}\Sigma_{\mathcal{D}_i\mathcal{D}_i}^{-1})y_{\mathcal{D}_i}\\
=\displaystyle\Sigma_{\mathcal{S}\mathcal{D}_i}\Sigma_{\mathcal{D}_i\mathcal{D}_i}^{-1}y_{\mathcal{D}_i}+\Sigma_{\mathcal{S}\mathcal{D}_i}\Sigma_{\mathcal{D}_i\mathcal{D}_i}^{-1}\Sigma_{\mathcal{D}_i\mathcal{S}}\Sigma_{\mathcal{S}\mathcal{S}|\mathcal{D}_i}^{-1}\Sigma_{\mathcal{S}\mathcal{D}_i}\Sigma_{\mathcal{D}_i\mathcal{D}_i}^{-1}y_{\mathcal{D}_i}\\
=\displaystyle\omega_{\mathcal{S}|\mathcal{D}_i}+\Phi_{\mathcal{S}\mathcal{S}|\mathcal{D}_i}(\Sigma_{\mathcal{S}\mathcal{S}}-\Sigma_{\mathcal{S}\mathcal{D}_i}\Sigma_{\mathcal{D}_i\mathcal{D}_i}^{-1}\Sigma_{\mathcal{D}_i\mathcal{S}})^{-1}\omega_{\mathcal{S}|\mathcal{D}_i}\\
=\displaystyle\omega_{\mathcal{S}|\mathcal{D}_i}+\Phi_{\mathcal{S}\mathcal{S}|\mathcal{D}_i}(\Sigma_{\mathcal{S}\mathcal{S}}-\Phi_{\mathcal{S}\mathcal{S}|\mathcal{D}_i})^{-1}\omega_{\mathcal{S}|\mathcal{D}_i}\\
=\displaystyle\Phi_{\mathcal{S}\mathcal{S}|\mathcal{D}_i}(\Phi_{\mathcal{S}\mathcal{S}|\mathcal{D}_i}^{-1}+(\Sigma_{\mathcal{S}\mathcal{S}}-\Phi_{\mathcal{S}\mathcal{S}|\mathcal{D}_i})^{-1})\omega_{\mathcal{S}|\mathcal{D}_i}\\
=\displaystyle\Phi_{\mathcal{S}\mathcal{S}|\mathcal{D}_i}(\Sigma_{\mathcal{S}\mathcal{S}}-\Phi_{\mathcal{S}\mathcal{S}|\mathcal{D}_i})^{-1}\\
\qquad\qquad\displaystyle((\Sigma_{\mathcal{S}\mathcal{S}}-\Phi_{\mathcal{S}\mathcal{S}|\mathcal{D}_i})\Phi_{\mathcal{S}\mathcal{S}|\mathcal{D}_i}^{-1}+I)\omega_{\mathcal{S}|\mathcal{D}_i}\\
=\displaystyle\Phi_{\mathcal{S}\mathcal{S}|\mathcal{D}_i}(\Sigma_{\mathcal{S}\mathcal{S}}-\Phi_{\mathcal{S}\mathcal{S}|\mathcal{D}_i})^{-1}\Sigma_{\mathcal{S}\mathcal{S}}\Phi_{\mathcal{S}\mathcal{S}|\mathcal{D}_i}^{-1}\omega_{\mathcal{S}|\mathcal{D}_i}\\
=\displaystyle(\Sigma_{\mathcal{S}\mathcal{S}}\Phi_{\mathcal{S}\mathcal{S}|\mathcal{D}_i}^{-1}-I)^{-1}\Sigma_{\mathcal{S}\mathcal{S}}\Phi_{\mathcal{S}\mathcal{S}|\mathcal{D}_i}^{-1}\omega_{\mathcal{S}|\mathcal{D}_i}\\
=\displaystyle(\Phi_{\mathcal{S}\mathcal{S}|\mathcal{D}_i}^{-1}-\Sigma_{\mathcal{S}\mathcal{S}}^{-1})^{-1}\Phi_{\mathcal{S}\mathcal{S}|\mathcal{D}_i}^{-1}\omega_{\mathcal{S}|\mathcal{D}_i}\\
=\displaystyle\Psi_{\mathcal{S}\mathcal{S}|\mathcal{D}_i}\Phi_{\mathcal{S}\mathcal{S}|\mathcal{D}_i}^{-1}\omega_{\mathcal{S}|\mathcal{D}_i}
\end{array}
\end{equation*}
where the second equality follows from the matrix inverse lemma on $\Sigma_{\mathcal{D}_i\mathcal{D}_i|\mathcal{S}}^{-1} =\Sigma_{\mathcal{D}_i\mathcal{D}_i}^{-1}+\Sigma_{\mathcal{D}_i\mathcal{D}_i}^{-1}\Sigma_{\mathcal{D}_i\mathcal{S}}\Sigma_{\mathcal{S}\mathcal{S}|\mathcal{D}_i}^{-1}\Sigma_{\mathcal{S}\mathcal{D}_i}\Sigma_{\mathcal{D}_i\mathcal{D}_i}^{-1}$.
As a result, $\Psi_{\mathcal{S}\mathcal{S}|\mathcal{D}_i}^{-1}\nu_{\mathcal{S}|\mathcal{D}_i} = \Phi_{\mathcal{S}\mathcal{S}|\mathcal{D}_i}^{-1}\omega_{\mathcal{S}|\mathcal{D}_i}$.
%
So,~\eqref{eqn:ppl} follows.
%
%
%
%
\section{Proof of Theorem~\ref{thm1}}\label{Apx.err}
%
The following lemma is necessary for deriving our main result here:
\begin{lem}
Define $\sigma_{xx'}$ using a squared exponential covariance function.
Then, every covariance component $\sigma_{xx'}$ in $\Sigma_{\mathcal{S}'_{t}\mathcal{D}_{it'}}$, $\Sigma_{\mathcal{S}\mathcal{S}}$, $\Sigma_{\mathcal{S}'_{t}\mathcal{S}}$, and $\Sigma_{\mathcal{D}_{it'}\mathcal{S}}$ satisfies
$(\sigma_{xx'}-\sigma_{ss'})^2\leq{3e^{-1}\sigma_s^4}(\|\Lambda^{-1}(x-s)\|^2 + \|\Lambda^{-1}(x'-s')\|^2)$
for all $x,x',s,s'\in\mathcal{X}$.
\label{secf}
\end{lem}
\emph{Proof.} Since every covariance component $\sigma_{xx'}$ in $\Sigma_{\mathcal{S}'_{t}\mathcal{D}_{it'}}$, $\Sigma_{\mathcal{S}\mathcal{S}}$, $\Sigma_{\mathcal{S}'_{t}\mathcal{S}}$, and $\Sigma_{\mathcal{D}_{it'}\mathcal{S}}$ does not involve the noise variance $\sigma^2_n$, it follows from~\eqref{secf1} that
$$
\begin{array}{rcl}
\sigma_{xx'} \hspace{-2.4mm}
&=&\hspace{-2.4mm}\displaystyle \sigma_s^2\exp\left(-\left\|\frac{{\Lambda}^{-1}({x} - {x}')}{\sqrt{2}}\right\|^2\right)\vspace{0.5mm}\\
&=&\hspace{-2.4mm}\displaystyle \sigma_s^2 k\left(\left\|\frac{{\Lambda}^{-1}({x} - {x}')}{\sqrt{2}}\right\|\right)
\end{array}
$$
where $k(a)\triangleq\exp(-a^2)$.
Then,
$$
\hspace{-1.7mm}
\begin{array}{l}
\displaystyle(\sigma_{xx'}-\sigma_{ss'})^2\\
\displaystyle = \sigma_s^4 \left\{k\left(\left\|\frac{{\Lambda}^{-1}({x} - {x}')}{\sqrt{2}}\right\|\right) - k\left(\left\|\frac{{\Lambda}^{-1}({s} - {s}')}{\sqrt{2}}\right\|\right)\right\}^2\vspace{1mm}\\
\displaystyle = 0.5\sigma_s^4 k'(\xi)^2(\|\Lambda^{-1}({x} - {x}')\| - \|\Lambda^{-1}({s} - {s}')\|)^2\vspace{1mm}\\
\displaystyle\leq e^{-1}\sigma_s^4 (\|\Lambda^{-1}(x-s)\| + \|\Lambda^{-1}(x'-s')\|)^2\vspace{1mm}\\
\displaystyle\leq e^{-1}\sigma_s^4(\|\Lambda^{-1}(x-s)\| + \|\Lambda^{-1}(x'-s')\|)^2\vspace{1mm}\\
\displaystyle\leq{3e^{-1}\sigma_s^4}(\|\Lambda^{-1}(x-s)\|^2 + \|\Lambda^{-1}(x'-s')\|^2)
\end{array}
$$
where the second equality is due to mean value theorem such that $k'(\xi)$ is the first-order derivative of $k$ evaluated at some $\xi\in(\|\Lambda^{-1}(s-s')\|/\sqrt{2},\|\Lambda^{-1}(x-x')\|/\sqrt{2})$ without loss of generality,
the first inequality follows from the fact that $k'(a)$ is maximized at $a=-1/\sqrt{2}$ and hence $k'(\xi)\leq k'(-1/\sqrt{2})=\sqrt{2/e}$,
and the second inequality is due to triangle inequality (i.e., $\|\Lambda^{-1}(x-x')\|\leq\|\Lambda^{-1}(x-s)\|+\|\Lambda^{-1}(s-s')\|+\|\Lambda^{-1}(s'-x')\|$).$\quad_\Box$

Supposing each subset $\mathcal{D}_{is}$ ($\mathcal{S}'_s$) contains $T$ ($T'$) locations\footnote{If the subset sizes differ, then ``virtual'' locations are added to each subset to make all subsets to be of the same size as $T\triangleq\arg\max_{s\in\mathcal{S}} |\mathcal{D}_{is}|$ ($T'\triangleq\arg\max_{s\in\mathcal{S}} |\mathcal{S}'_{s}|$). The virtual locations added to $\mathcal{D}_{is}$ ($\mathcal{S}'_{s}$) are chosen as $s\in\mathcal{S}$ so that they do not induce additional errors but will loosen the bound.}, select one location from each subset to form a new subset $\mathcal{D}_{it'}\triangleq\{x_{it's} \}_{s\in\mathcal{S}}$ ($\mathcal{S}'_t\triangleq\{x'_{ts} \}_{s\in\mathcal{S}}$) of $|\mathcal{S}|$ locations for $t'=1$ ($t=1$) and repeat this for $t'=2,\ldots,T$ ($t=2,\ldots,T'$).
Then, $\mathcal{D}_{i}=\bigcup^T_{t'=1}\mathcal{D}_{it'}$ and $\mathcal{S}'=\bigcup^{T'}_{t=1}\mathcal{S}'_{t}$.
It follows that $\Sigma_{\mathcal{S}'\mathcal{S}} = [\Sigma_{\mathcal{S}'_{t}\mathcal{S}}]_{t=1,\ldots,T'}$, $\Sigma_{\mathcal{S}\mathcal{D}_{i}} = [\Sigma_{\mathcal{S}\mathcal{D}_{it'}}]_{t'=1,\ldots,T}$, and $\Sigma_{\mathcal{S}'\mathcal{D}_{i}} = [\Sigma_{\mathcal{S}'_{t}\mathcal{D}_{it'}}]_{t=1,\ldots,T',t'=1,\ldots,T}$.

Using the definition of Frobenius norm followed by the subadditivity of a square root function,
\begin{equation}
\hspace{-1.7mm}
\begin{array}{l}
\displaystyle ||\Sigma_{\mathcal{S}'\mathcal{D}_i} - \Sigma_{\mathcal{S}'\mathcal{S}}\Sigma_{\mathcal{S}\mathcal{S}}^{-1}\Sigma_{\mathcal{S}\mathcal{D}_i}||_F\\
= \displaystyle ||\Sigma_{\mathcal{S}'\mathcal{D}_i|\mathcal{S}}||_F\\
= \displaystyle \sqrt{\sum^{T'}_{t=1} \sum^{T}_{t'=1} ||\Sigma_{\mathcal{S}'_{t}\mathcal{D}_{it'}|\mathcal{S}}||^2_F}\\
\leq \displaystyle \sum^{T'}_{t=1} \sum^{T}_{t'=1} ||\Sigma_{\mathcal{S}'_{t}\mathcal{D}_{it'}|\mathcal{S}}||_F\ .
\end{array}
\label{tele}
\end{equation}
Let $A_{\mathcal{S}'_{t}\mathcal{D}_{it'}}\triangleq\Sigma_{\mathcal{S}'_{t}\mathcal{D}_{it'}}-\Sigma_{\mathcal{S}\mathcal{S}}$, $B_{\mathcal{S}'_t\mathcal{S}}\triangleq\Sigma_{\mathcal{S}'_t\mathcal{S}}-\Sigma_{\mathcal{S}\mathcal{S}}$, and
$C_{\mathcal{D}_{it'}\mathcal{S}}\triangleq\Sigma_{\mathcal{D}_{it'}\mathcal{S}}-\Sigma_{\mathcal{S}\mathcal{S}}$. Then,
\begin{equation}
\hspace{-1.7mm}
\begin{array}{l}
\displaystyle ||\Sigma_{\mathcal{S}'_{t}\mathcal{D}_{it'}|\mathcal{S}}||_F\\
= \displaystyle  ||\Sigma_{\mathcal{S}'_{t}\mathcal{D}_{it'}} - \Sigma_{\mathcal{S}'_t\mathcal{S}}\Sigma_{\mathcal{S}\mathcal{S}}^{-1}\Sigma_{\mathcal{S}\mathcal{D}_{it'}}||_F\\
= \displaystyle || \Sigma_{\mathcal{S}\mathcal{S}}+A_{\mathcal{S}'_{t}\mathcal{D}_{it'}}\ -\\\displaystyle\quad\ \ (\Sigma_{\mathcal{S}\mathcal{S}}+B_{\mathcal{S}'_t\mathcal{S}})\Sigma_{\mathcal{S}\mathcal{S}}^{-1}(\Sigma_{\mathcal{S}\mathcal{S}}+C_{\mathcal{D}_{it'}\mathcal{S}})^{\top}||_F\\
= \displaystyle || \Sigma_{\mathcal{S}\mathcal{S}}+A_{\mathcal{S}'_{t}\mathcal{D}_{it'}} -\Sigma_{\mathcal{S}\mathcal{S}}^{\top}-C_{\mathcal{D}_{it'}\mathcal{S}}^{\top}
-B_{\mathcal{S}'_t\mathcal{S}}\ -\\
\displaystyle\quad\ \ B_{\mathcal{S}'_t\mathcal{S}}\Sigma_{\mathcal{S}\mathcal{S}}^{-1}C_{\mathcal{D}_{it'}\mathcal{S}}^{\top}||_F\\
\leq \displaystyle || A_{\mathcal{S}'_{t}\mathcal{D}_{it'}}||_F
+||B_{\mathcal{S}'_t\mathcal{S}}||_F
+||C_{\mathcal{D}_{it'}\mathcal{S}}||_F
\ +\\
\quad\displaystyle ||B_{\mathcal{S}'_t\mathcal{S}}||_F
||C_{\mathcal{D}_{it'}\mathcal{S}}||_F
||\Sigma_{\mathcal{S}\mathcal{S}}^{-1}||_F
\end{array}
\label{tele2}
\end{equation}
where the inequality is due to the subadditivity and submultiplicativity of the matrix norm.

Let $\epsilon_{\mathcal{S}'_{t}}\triangleq(1/|\mathcal{S}|)\sum_{x\in\mathcal{S}'_{t}}||\Lambda^{-1}(x-c(x))||^2$ and
$\epsilon_{\mathcal{D}_{it'}}\triangleq(1/|\mathcal{S}|)\sum_{x\in\mathcal{D}_{it'}}||\Lambda^{-1}(x-c(x))||^2$. Then,
\begin{equation}
\hspace{-1.7mm}
\begin{array}{l}
 \displaystyle || A_{\mathcal{S}'_{t}\mathcal{D}_{it'}}||^2_F\\
= \displaystyle ||\Sigma_{\mathcal{S}'_{t}\mathcal{D}_{it'}}-\Sigma_{\mathcal{S}\mathcal{S}}||^2_F\\
= \displaystyle\sum_{s,s'\in\mathcal{S}} (\sigma_{x'_{ts}x_{it's'}}-\sigma_{ss'})^2\\
\leq 3e^{-1}\sigma^4_s \displaystyle\sum_{s,s'\in\mathcal{S}}\left(||\Lambda^{-1}(x'_{ts}-s)||^2 + ||\Lambda^{-1}(x_{it's'}-s')||^2\right)\\
= 3e^{-1}\sigma^4_s|\mathcal{S}|\displaystyle\Bigg(\sum_{s\in\mathcal{S}}||\Lambda^{-1}(x'_{ts}-s)||^2\ + \\
\displaystyle\qquad\qquad\qquad\ \sum_{s'\in\mathcal{S}}||\Lambda^{-1}(x_{it's'}-s')||^2\Bigg)\\
= 3e^{-1}\sigma^4_s|\mathcal{S}|^2\displaystyle(\epsilon_{\mathcal{S}'_{t}} + \epsilon_{\mathcal{D}_{it'}})
\end{array}
\label{tele3}
\end{equation}
since $\epsilon_{\mathcal{S}'_{t}}=(1/|\mathcal{S}|)\sum_{s\in\mathcal{S}}||\Lambda^{-1}(x'_{ts}-s)||^2$
and $\epsilon_{\mathcal{D}_{it'}}=(1/|\mathcal{S}|)\sum_{s'\in\mathcal{S}}||\Lambda^{-1}(x_{it's'}-s')||^2$.
The inequality is due to Lemma~\ref{secf}.
\begin{equation}
\begin{array}{l}
 \displaystyle || B_{\mathcal{S}'_{t}\mathcal{S}}||^2_F\\
= \displaystyle ||\Sigma_{\mathcal{S}'_{t}\mathcal{S}}-\Sigma_{\mathcal{S}\mathcal{S}}||^2_F\\
= \displaystyle\sum_{s,s'\in\mathcal{S}} (\sigma_{x'_{ts}s'}-\sigma_{ss'})^2\\
\leq 3e^{-1}\sigma^4_s\displaystyle\sum_{s,s'\in\mathcal{S}}\left(||\Lambda^{-1}(x'_{ts}-s)||^2 + ||\Lambda^{-1}(s'-s')||^2\right)\\
= 3e^{-1}\sigma^4_s|\mathcal{S}|\displaystyle\sum_{s\in\mathcal{S}}||\Lambda^{-1}(x'_{ts}-s)||^2\\
= 3e^{-1}\sigma^4_s|\mathcal{S}|^2\displaystyle\epsilon_{\mathcal{S}'_{t}}
\end{array}
\label{tele4}
\end{equation}
such that the inequality is due to Lemma~\ref{secf}.
\begin{equation}
\begin{array}{l}
 \displaystyle || C_{\mathcal{D}_{it'}\mathcal{S}}||^2_F\\
= \displaystyle ||\Sigma_{\mathcal{D}_{it'}\mathcal{S}}-\Sigma_{\mathcal{S}\mathcal{S}}||^2_F\\
= \displaystyle\sum_{s,s'\in\mathcal{S}} (\sigma_{x_{it's}s'}-\sigma_{ss'})^2\\
\leq 3e^{-1}\sigma^4_s\displaystyle\sum_{s,s'\in\mathcal{S}}\left(||\Lambda^{-1}(x_{it's}-s)||^2 + ||\Lambda^{-1}(s'-s')||^2\right)\\
= 3e^{-1}\sigma^4_s|\mathcal{S}|\displaystyle\sum_{s\in\mathcal{S}}||\Lambda^{-1}(x_{it's}-s)||^2\\
= 3e^{-1}\sigma^4_s|\mathcal{S}|^2\displaystyle\epsilon_{\mathcal{D}_{it'}}
\end{array}
\label{tele5}
\end{equation}
such that the inequality is due to Lemma~\ref{secf}.

By substituting~\eqref{tele3},~\eqref{tele4}, and~\eqref{tele5} into~\eqref{tele2},
\begin{equation}
\hspace{-1.7mm}
\begin{array}{l}
\displaystyle ||\Sigma_{\mathcal{S}'_{t}\mathcal{D}_{it'}|\mathcal{S}}||_F\\
\leq \displaystyle \sqrt{3e^{-1}\sigma^4_s|\mathcal{S}|^2\displaystyle(\epsilon_{\mathcal{S}'_{t}} + \epsilon_{\mathcal{D}_{it'}})}
+\sqrt{3e^{-1}\sigma^4_s|\mathcal{S}|^2\displaystyle\epsilon_{\mathcal{S}'_{t}}}\
+\\
\quad\displaystyle\sqrt{3e^{-1}\sigma^4_s|\mathcal{S}|^2\displaystyle\epsilon_{\mathcal{D}_{it'}}}
\ +\\
\quad\displaystyle \sqrt{3e^{-1}\sigma^4_s|\mathcal{S}|^2\displaystyle\epsilon_{\mathcal{S}'_{t}}}
\sqrt{3e^{-1}\sigma^4_s|\mathcal{S}|^2\displaystyle\epsilon_{\mathcal{D}_{it'}}}
||\Sigma_{\mathcal{S}\mathcal{S}}^{-1}||_F\\
=\displaystyle \sqrt{3/e}\sigma^2_s|\mathcal{S}|\Big(\sqrt{\epsilon_{\mathcal{S}'_{t}} + \epsilon_{\mathcal{D}_{it'}}}+\sqrt{\epsilon_{\mathcal{S}'_{t}}}+\sqrt{\epsilon_{\mathcal{D}_{it'}}}\ +\\
\displaystyle\qquad\qquad\qquad\ \ \sigma^2_s||\Sigma_{\mathcal{S}\mathcal{S}}^{-1}||_F|\mathcal{S}|\sqrt{3\epsilon_{\mathcal{S}'_{t}}\epsilon_{\mathcal{D}_{it'}}/e}\Big)\ .
\end{array}
\label{tele6}
\end{equation}
By substituting~\eqref{tele6} into~\eqref{tele},
\begin{equation*}
\hspace{-1.7mm}
\begin{array}{l}
\displaystyle ||\Sigma_{\mathcal{S}'\mathcal{D}_i} - \Sigma_{\mathcal{S}'\mathcal{S}}\Sigma_{\mathcal{S}\mathcal{S}}^{-1}\Sigma_{\mathcal{S}\mathcal{D}_i}||_F\\
\leq \displaystyle \sqrt{3/e}\sigma^2_s|\mathcal{S}|\sum^{T'}_{t=1} \sum^{T}_{t'=1} \Big(\sqrt{\epsilon_{\mathcal{S}'_{t}} + \epsilon_{\mathcal{D}_{it'}}}+\sqrt{\epsilon_{\mathcal{S}'_{t}}}+\sqrt{\epsilon_{\mathcal{D}_{it'}}}\ +\\
\displaystyle\qquad\qquad\qquad\qquad\qquad \sigma^2_s||\Sigma_{\mathcal{S}\mathcal{S}}^{-1}||_F|\mathcal{S}|\sqrt{3\epsilon_{\mathcal{S}'_{t}}\epsilon_{\mathcal{D}_{it'}}/e}\Big)\\
\leq \displaystyle \sqrt{3/e}\sigma^2_s|\mathcal{S}| \Bigg(\sqrt{TT'\sum^{T'}_{t=1} \sum^{T}_{t'=1} (\epsilon_{\mathcal{S}'_{t}} + \epsilon_{\mathcal{D}_{it'}})}\ +\\
\displaystyle\qquad\qquad\quad\ \sqrt{TT'\sum^{T'}_{t=1} \sum^{T}_{t'=1} \epsilon_{\mathcal{S}'_{t}}} + \sqrt{TT'\sum^{T'}_{t=1} \sum^{T}_{t'=1} \epsilon_{\mathcal{D}_{it'}}}\ +\\
\displaystyle\qquad\qquad\quad\ \sigma^2_s||\Sigma_{\mathcal{S}\mathcal{S}}^{-1}||_F|\mathcal{S}|\sqrt{TT'(3/e)\sum^{T'}_{t=1} \sum^{T}_{t'=1} \epsilon_{\mathcal{S}'_{t}}\epsilon_{\mathcal{D}_{it'}}}\Bigg)\\
= \displaystyle \sqrt{3/e}\sigma^2_s|\mathcal{S}|\Bigg(\sqrt{TT'\left(T\sum^{T'}_{t=1}  \epsilon_{\mathcal{S}'_{t}} + T'\sum^{T}_{t'=1}\epsilon_{\mathcal{D}_{it'}}\right)}\ +\\
\displaystyle\qquad\qquad\qquad\ \ \sqrt{T^2T'\sum^{T'}_{t=1}  \epsilon_{\mathcal{S}'_{t}}} + \sqrt{TT'^2 \sum^{T}_{t'=1} \epsilon_{\mathcal{D}_{it'}}}\ +\\
\displaystyle\qquad\qquad\quad \sigma^2_s||\Sigma_{\mathcal{S}\mathcal{S}}^{-1}||_F|\mathcal{S}|\sqrt{TT'(3/e)\sum^{T'}_{t=1} \epsilon_{\mathcal{S}'_{t}}\sum^{T}_{t'=1} \epsilon_{\mathcal{D}_{it'}}}\Bigg)\\
= \displaystyle \sqrt{3/e}\sigma^2_s|\mathcal{S}|TT' \Big(\sqrt{\epsilon_{\mathcal{S}'} + \epsilon_{\mathcal{D}_{i}}} +\sqrt{\epsilon_{\mathcal{S}'}}+\sqrt{\epsilon_{\mathcal{D}_{i}}}\ +\\
\displaystyle\qquad\qquad\qquad\qquad \ \sigma^2_s||\Sigma_{\mathcal{S}\mathcal{S}}^{-1}||_F|\mathcal{S}|\sqrt{3\epsilon_{\mathcal{S}'}\epsilon_{\mathcal{D}_{i}}/e}\Big)
\end{array}
\end{equation*}
such that the second inequality follows from
$$
\sum^T_{t=1}\sqrt{a_t}\leq \sqrt{T\sum^T_{t=1}a_t}
$$
which can be obtained by applying Jensen's inequality to the concave square root function.
The last equality is due to $\epsilon_{\mathcal{S}'} = (1/T')\sum^{T'}_{t=1}\epsilon_{\mathcal{S}'_t}$ and
$\epsilon_{\mathcal{D}_{i}} = (1/T)\sum^{T}_{t'=1}\epsilon_{\mathcal{D}_{it'}}$.
\section{GP-DDF/GP-DDF$^+$ Algorithm with Agent-Centric Support Sets based on Lazy Transfer Learning}
\label{runforest}
Refer to Algorithm~\ref{alg:sendrecv} below.
\begin{algorithm}\label{alg:sendrecv}
\begin{small}
\DontPrintSemicolon
\If{\emph{agent $i$ transits from local area with support set $\mathcal{S}$ to local area with support set $\mathcal{S}'$}} {
\tcc{Information sharing mechanism}
\tcc{Leaving local area with $\mathcal{S}$}
\If{\emph{other agents are in local area with support set $\mathcal{S}$}} {
    Construct and send local summary $(\nu_{\mathcal{S}|\mathcal{D}_i},\hspace{-0.8mm}\Psi_{\mathcal{S}\mathcal{S}|\mathcal{D}_i})$ to an agent in this area who assimilates it with its own local summary using~\eqref{eqn:globalsum};\;
    Delete local summary $(\nu_{\mathcal{S}|\mathcal{D}_i},\Psi_{\mathcal{S}\mathcal{S}|\mathcal{D}_i})$;
}
\Else {
    Backup local summary $(\nu_{\mathcal{S}|\mathcal{D}_i},\Psi_{\mathcal{S}\mathcal{S}|\mathcal{D}_i})$;
}
\tcc{Entering local area with $\mathcal{S}'$}
\If{\emph{other agents are in local area with support set $\mathcal{S}'$}} {
    Get support set $\mathcal{S}'$ from an agent in this area;\;
}
\Else{
    \If{\emph{some agent $j$ in the team stores a backup of local summary based on support set $\mathcal{S}'$}} {
            Retrieve and remove this backup of local summary based on $\mathcal{S}'$ from agent $j$;\;
    }
    \Else {
        Construct support set $\mathcal{S}'$;\;
    }
}
}
\If{\emph{agent $i$ has to predict the phenomenon}} {
\If{\emph{data $(\mathcal{D}'_i, y_{\mathcal{D}'_i})$ is available from local area with support set $\mathcal{S}'$}} {Construct local summary $(\nu_{\mathcal{S}'|\mathcal{D}'_i},\Psi_{\mathcal{S}'\mathcal{S}'|\mathcal{D}'_i})$ by~\eqref{eqn:localsum};}
Exchange local summary with every agent $j\neq i$;\;
\ForEach{\emph{agent $j\neq i$ in local area with support set $\mathcal{S}''\neq \mathcal{S}'$}}
{
\tcc{Transfer learning mechanism}
Derive local summary $(\nu_{\mathcal{S}'|\mathcal{D}_j},\Psi_{\mathcal{S}'\mathcal{S}'|\mathcal{D}_j})$ based on $\mathcal{S}'$ approximately from received local summary $(\nu_{\mathcal{S}''|\mathcal{D}_j},\Psi_{\mathcal{S}''\mathcal{S}''|\mathcal{D}_j})$ based on $\mathcal{S}''$ using  transfer learning mechanism in Algorithm~\ref{alg:local} (Section~\ref{sect:tl});}
Compute global summary $(\dot{\nu}_{\mathcal{S}'},\dot{\Psi}_{\mathcal{S}'\mathcal{S}'})$ by~\eqref{eqn:globalsum} using local summaries $(\nu_{\mathcal{S}'|\mathcal{D}'_i},\Psi_{\mathcal{S}'\mathcal{S}'|\mathcal{D}'_i})$ and $(\nu_{\mathcal{S}'|\mathcal{D}_j},\Psi_{\mathcal{S}'\mathcal{S}'|\mathcal{D}_j})$ of every agent $j\neq i$;\;
Run GP-DDF~\eqref{eqn:pred} or GP-DDF$^+$~\eqref{supercareless};\;
}
\end{small}
\caption{GP-DDF/GP-DDF$^+$ with agent-centric support sets based on lazy transfer learning for agent $i$}
\end{algorithm}
\section{Real-World Plankton Density Phenomenon}
\label{plankton}
The MODIS plankton density dataset (Fig.~\ref{fig:plankton}) is bounded within lat. $30$-$31$N and lon. $245.36$-$246.11$E (i.e., off the west coast of USA) with a data size of $4941$. The domain of this phenomenon is discretized into a $61 \times 81$ grid of locations that are associated with log-chlorophyll-a measurements in mg/m$^3$. It is partitioned into $K=16$ disjoint local areas of size about $15$ by $20$, each of which is assigned $N/K$ mobile sensing agents.
The $N/K$ agents in every local area then move together in a pre-defined lawnmower pattern from one local area to the next adjacent one such that they visit all the $K=16$ local areas exactly twice to gather data/observations from this phenomenon and end in the same local area initially assigned to them.
Whenever the $N/K$ agents transit into the next local area, they will move randomly within to gather the local data/observations; the results are averaged over $30$ runs.
\begin{figure}[h]
\centering
\includegraphics[width=5.5cm]{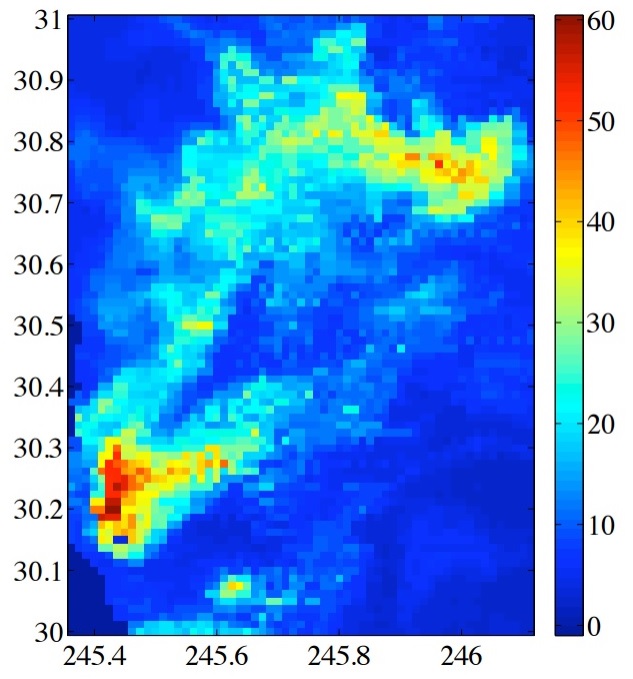}
\caption{Plankton density phenomenon bounded within lat. $30$-$31$N and lon. $245.36$-$246.11$E.}
\label{fig:plankton}
\end{figure}

The performance of our \emph{GP-DDF and GP-DDF$^+$ algorithms with agent-centric support sets} (respectively, GP-DDF-ASS and GP-DDF$^+$-ASS),
each of which is of size $81$ and uniformly distributed\cref{boo} over a different local area of the plankton density phenomenon,
are compared against that of the local GPs method 
and state-of-the-art GP-DDF and GP-DDF$^+$ \cite{LowTASE15} with a common support set of size $81$ uniformly distributed over the entire plankton density phenomenon and known to all agents.\vspace{0.5mm}

\noindent
{\bf Predictive Performance.} Fig.~\ref{fig:rmse2}a shows results of decreasing RMSE achieved by tested algorithms with an increasing total number of observations for $N=32$ agents. The observations and analysis are similar to that reported in Section~\ref{sect:expt} (specifically, under `Predictive Performance').
It can also be observed that the performance gap between GP-DDF-ASS and GP-DDF$^+$-ASS appears to be smaller than that for the indoor lighting quality and  temperature phenomenon shown in Figs.~\ref{fig:rmse}a and~\ref{fig:rmse}c, respectively: Compared to the indoor lighting quality (temperature phenomenon), the plankton density phenomenon has a relatively larger length-scale (much smaller domain size and consequently closer agent-centric support sets), thereby making transfer learning more effective, which agrees with the observation in our theoretical analysis for Theorem~\ref{thm1} (Section~\ref{guarantee}), and reducing the performance advantage of GP-DDF$^+$-ASS over GP-DDF-ASS in exploiting local data.\vspace{0.5mm}
\begin{figure}
\centering
\begin{tabular}{ccc}
  \hspace{-3mm}
  \includegraphics[scale=0.15]{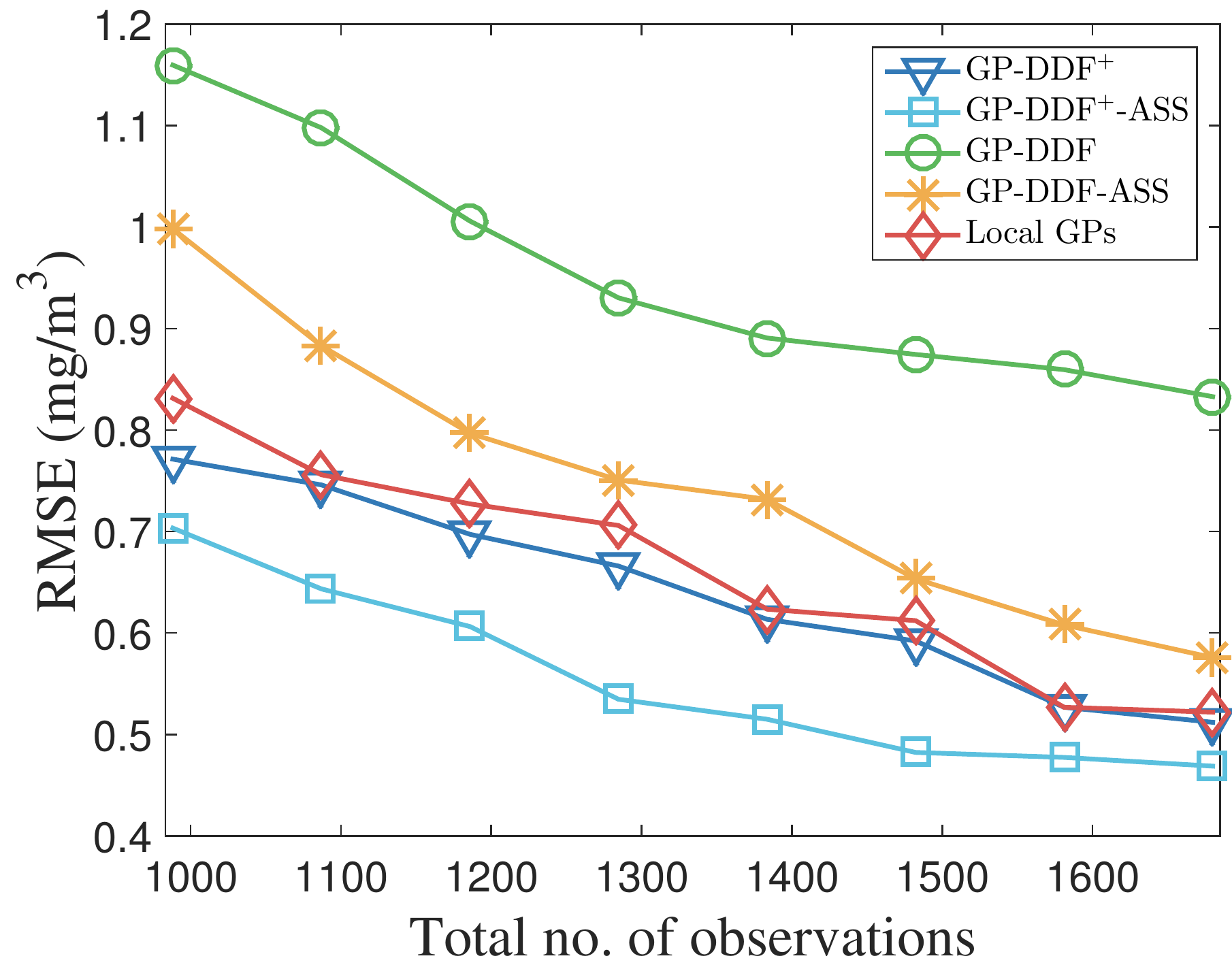} & \hspace{-4mm}
  \includegraphics[scale=0.15]{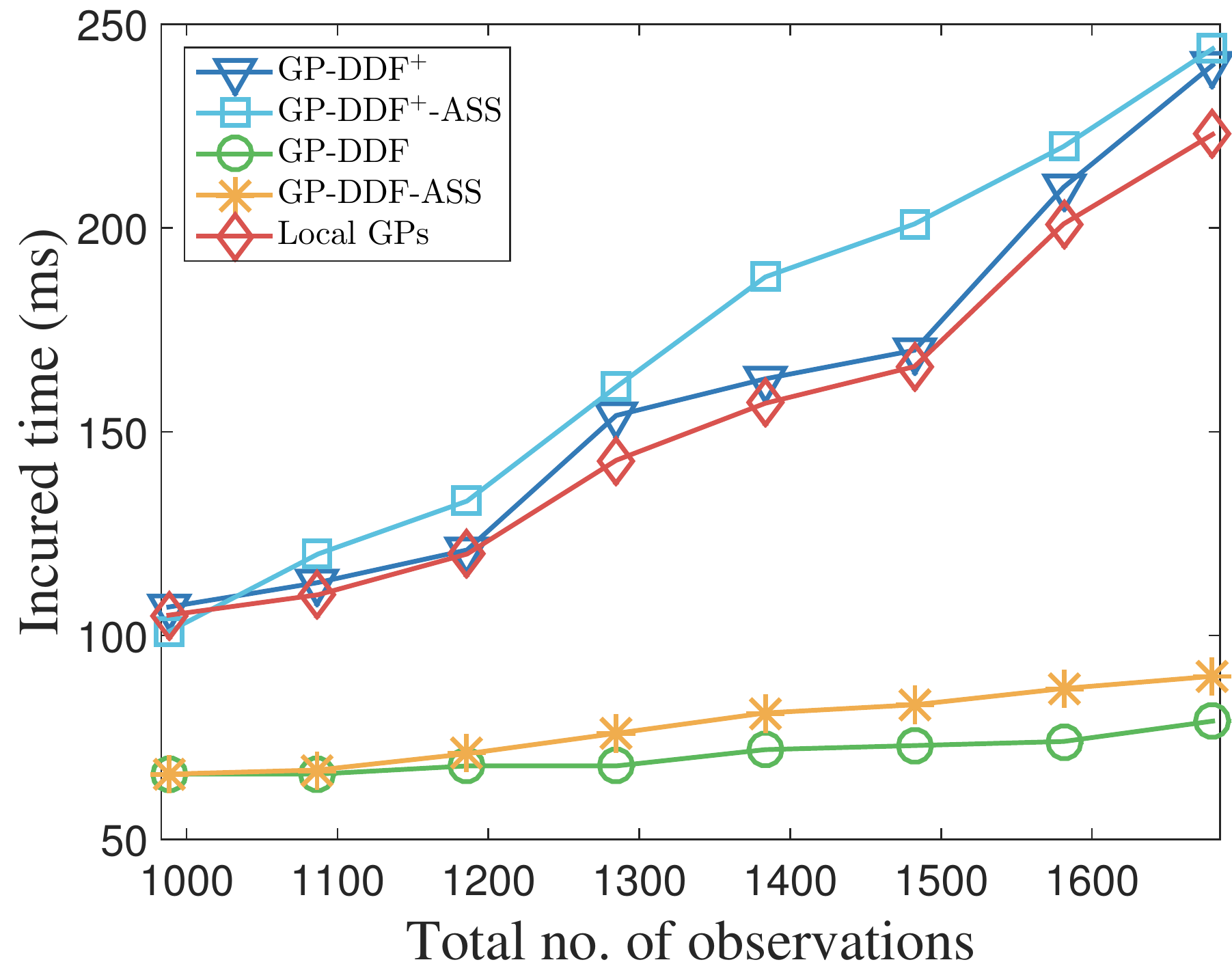} & \hspace{-4mm}
  \includegraphics[scale=0.15]{total_p}\\
  \hspace{-3mm}
  (a)  & \hspace{-4mm}
  (b) & \hspace{-4mm}
  (c) \vspace{-2mm}
\end{tabular}
\caption{Graphs of (a) RMSE and (b) total incurred time vs. total no. of observations, and (c) graphs of total incurred time vs. no. of agents achieved by tested algorithms for plankton density phenomenon.
}\label{fig:rmse2}
\end{figure}

\noindent
{\bf Time Efficiency.} Fig.~\ref{fig:rmse2}b shows results of increasing total time incurred by tested algorithms with an increasing total number of observations for $N=32$ agents. The experimental setup, observations, and analysis are again similar to that reported in Section~\ref{sect:expt} (specifically, under `Time Efficiency').\vspace{0.5mm}

\noindent
{\bf Scalability in the Number of Agents.} Fig.~\ref{fig:rmse2}c shows results of total time incurred by tested algorithms with an increasing number $N$ of agents to gather a total number of $1235$ observations.
%
%
It can be observed that the total time incurred by GP-DDF$^+$-ASS, GP-DDF$^+$, GP-DDF-ASS, and GP-DDF decrease with more agents, as explained in Section~\ref{notcareful}; recall further that they become more robust to agent failure with more agents assigned to every local area to reduce its risk of being empty and hence its likelihood of inducing a backup.
In addition, GP-DDF-ASS and GP-DDF$^+$-ASS, respectively,  incur only slightly more time than GP-DDF and GP-DDF$^+$ due to their information sharing mechanism described in Section~\ref{gpddfass} (specifically, the first if-then construct in Algorithm~\ref{alg:sendrecv} in Appendix~\ref{runforest}).
Note that the total time incurred by local GPs remains constant for $N\geq 16$ agents because a fixed number of about $77$ observations are gathered in each local area and used by one of the $N$ agents for prediction in that local area; when $N=8$, every agent has to perform prediction for $2$ of the $16$ local areas instead, hence incurring twice the amount of time.
Note that the local GPs method requires all the $77$ observations gathered in each local area by different agents to be communicated to the agent performing prediction in that local area.

%
%
%
\section{Indoor Lighting Quality}
\label{robot}
Refer to Fig.~\ref{fig:robot}.
\begin{figure}[h]
\centering
\begin{tabular}{c}
  \includegraphics[width=8.4cm, height=6cm]{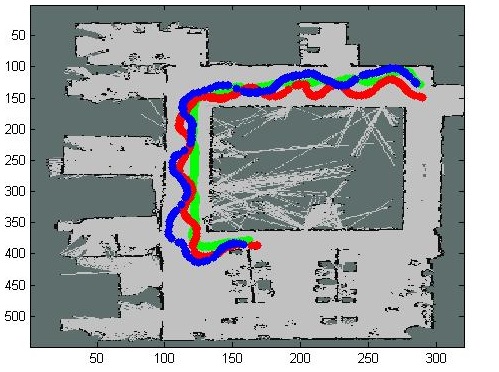}\vspace{-2mm}\\
  \hspace{-4mm} (a)\\
  \hspace{-4mm}\includegraphics[width=8cm, height=6cm]{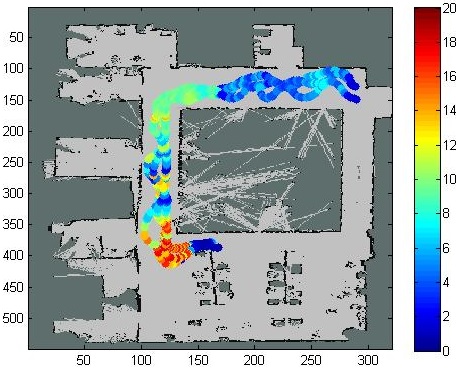}\vspace{-2mm}\\
   \hspace{-4mm}(b) \vspace{-2mm}
\end{tabular}
\caption{(a) Red, green, and blue trajectories of three Pioneer $3$-DX mobile robots in an office environment generated by AMCL package in ROS, along which (b) $1200$ observations of relative lighting level are gathered simultaneously by the robots at locations denoted by small colored circles.}
\label{fig:robot}
\end{figure}

\section{Real-World Temperature Phenomenon}
\label{temperature}
Refer to Fig.~\ref{fig:temperature}.
\begin{figure}[h]
\includegraphics[width=8.6cm]{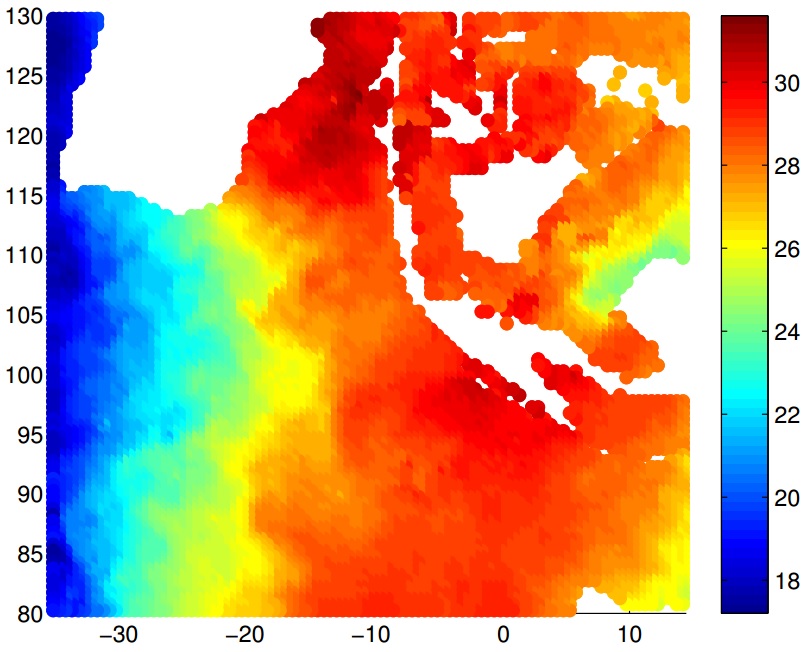}
\caption{Temperature phenomenon bounded within lat. $35.75$-$14.25$S and lon. $80.25$-$104.25$E in Dec. $2015$.}
\label{fig:temperature}
\end{figure}

\section{Hyperparameter Learning}
\label{hyper}
The hyperparameters of our GP-DDF-ASS and GP-DDF$^+$-ASS algorithms are learned by maximizing the sum of log-marginal likelihoods $\sum_{\mathcal{S}} \log p(y_\mathcal{D}|\mathcal{S})$ over the support set $\mathcal{S}$ of every different local area via gradient ascent with respect to a common set of signal variance, noise variance, and length-scale hyperparameters (Section~\ref{gpm}) where, as derived in \cite{candela05},
$$
\log p(y_\mathcal{D}|\mathcal{S}) = -0.5 (\log |\Xi_{\mathcal{D}\mathcal{D}|\mathcal{S}}|+y^{\top}_\mathcal{D}\Xi_{\mathcal{D}\mathcal{D}|\mathcal{S}}^{-1}y_\mathcal{D}+ |\mathcal{D}|\log (2\pi))
$$
such that
$\Xi_{\mathcal{D}\mathcal{D}|\mathcal{S}}\triangleq\Phi_{\mathcal{D}\mathcal{D}|\mathcal{S}}+\text{blockdiag}[\Sigma_{\mathcal{D}\mathcal{D}|\mathcal{S}}]+\sigma^2_n I$.
Note that these learned hyperparameters of our GP-DDF-ASS and GP-DDF$^+$-ASS algorithms correspond to the case where our proposed lazy transfer learning mechanism incurs minimal information loss, as explained in Appendix~\ref{toy}.
%
%
%
\fi
\end{document}